\documentclass{article}

\PassOptionsToPackage{numbers, compress}{natbib}
\usepackage[preprint]{style/neurips_2024}

\usepackage[utf8]{inputenc}
\usepackage[T1]{fontenc}
\usepackage{microtype}
\usepackage{subcaption}
\usepackage[dvipsnames]{xcolor}

\usepackage{caption}
\usepackage{graphicx}
\usepackage{pdflscape}

\usepackage{makecell}
\usepackage{booktabs}
\usepackage{tabularx}
\usepackage{array}

\usepackage[boxed]{algorithm2e}
\usepackage[noend]{algcompatible}

\usepackage{amsmath,amsfonts,bm, amssymb}
\usepackage{amsthm}
\usepackage{mathtools}
\usepackage{nicefrac}

\usepackage{setspace}
\usepackage{enumitem}
\usepackage[titles]{tocloft}
\usepackage{lipsum}
\usepackage[symbol]{footmisc}

\usepackage[pagebackref=true, hidelinks]{hyperref}
\usepackage{url}
\usepackage[noabbrev,nameinlink,capitalise,compress]{cleveref}
\usepackage{crossreftools} %
\usepackage{hypernat} %
\usepackage{nameref}
\renewcommand*{\backrefalt}[4]{\ifcase #1 No citations. \or Cited on page #2. \else Cited on pages #2. \fi}
\crefalias{inequality}{equation}
\crefalias{appendix}{section}
\crefalias{subappendix}{subsection}
\crefname{appendix}{}{}
\crefname{inequality}{Inequality}{Inequalities}
\crefname{property}{Property}{Properties}
\crefformat{section}{#2\S#1#3}
\crefformat{subappendix}{#2Appendix #1#3}
\creflabelformat{inequality}{#2\textup{(#1)}#3}
\newcommand{\mycaption}[2]{\caption[#1]{\textbf{#1} #2}}
\SetAlCapSkip{1em}

\let\oldaddcontentsline\addcontentsline
\newcommand{\stoptocentries}{\renewcommand{\addcontentsline}[3]{}}
\newcommand{\starttocentries}{\let\addcontentsline\oldaddcontentsline}
\newtheorem{proposition}{Proposition}

\newtheorem{definition}{Definition}

\theoremstyle{definition}

\newtheorem{atomic}{Atom}
\newtheorem{bond}{Bond}
\newtheorem{compound}{Compound}

\newtheorem{remark}{Remark}

\DeclareMathOperator*{\argmin}{arg\,min}

\newcommand{\defeq}{\vcentcolon=}
\newcommand{\el}{\mathcal{L}}
\newcommand{\data}{\mathcal{D}}

\newcommand{\softmax}{\operatorname{softmax}}

\newcommand{\half}{\tfrac{1}{2}}

\newcommand{\R}{\mathbb{R}}

\newcommand{\sign}{\operatorname{sign}}

\newcommand{\diag}{\operatorname{diag}}

\newcommand{\abs}[1]{\vert {#1} \vert}
\newcommand{\norm}[1]{\Vert {#1} \Vert}
\newcommand{\ip}[2]{\langle {#1}, \ {#2} \rangle}

\newcommand{\diff}{\mathrm{d}}
\newcommand{\idiff}{\,\diff}

\newcommand{\Expect}{\operatorname{\mathbb{E}}}

\def\eqref#1{equation~\ref{#1}}

\def\1{\bm{1}}

\def\vf{{\bm{f}}}
\def\vg{{\bm{g}}}
\def\vh{{\bm{h}}}

\def\vk{{\bm{k}}}

\def\vp{{\bm{p}}}
\def\vq{{\bm{q}}}

\def\vt{{\bm{t}}}

\def\vv{{\bm{v}}}
\def\vw{{\bm{w}}}
\def\vx{{\bm{x}}}
\def\vy{{\bm{y}}}
\def\vz{{\bm{z}}}

\def\mA{{\bm{A}}}
\def\mB{{\bm{B}}}
\def\mC{{\bm{C}}}

\def\mE{{\bm{E}}}

\def\mM{{\bm{M}}}

\def\mW{{\bm{W}}}

\DeclareMathAlphabet{\mathsfit}{\encodingdefault}{\sfdefault}{m}{sl}
\SetMathAlphabet{\mathsfit}{bold}{\encodingdefault}{\sfdefault}{bx}{n}

\newcommand{\out}{\mathrm{out}}
\newcommand{\inn}{\mathrm{in}}

\newcommand{\weights}{\mathcal{W}}
\newcommand{\inputs}{\mathcal{X}}
\newcommand{\outputs}{\mathcal{Y}}
\newcommand{\targets}{\mathcal{T}}

\newcommand{\for}{\mathsf{.forward}}
\newcommand{\mass}{\mathsf{.mass}}
\newcommand{\nor}{\mathsf{.norm}}

\newcommand{\lip}{\mathsf{.sensitivity}}

\newcommand{\Normalize}{\mathsf{normalize}}

\newcommand{\D}{\Delta}
\renewcommand{\d}{\partial}

\newcommand{\module}{\mathsf{M}}

\newcommand{\add}{\mathsf{Add}}
\newcommand{\mult}{\mathsf{Mul}}
\newcommand{\identity}{\mathsf{Identity}}
\newcommand{\layernorm}{\mathsf{LayerNorm}}
\newcommand{\residual}[2]{\mathsf{Res}_{#1}(#2)}
\newcommand{\relu}{\mathsf{ReLU}}
\newcommand{\gelu}{\mathsf{GELU}}
\newcommand{\scaledrelu}{\mathsf{ScaledReLU}}
\newcommand{\scaledgelu}{\mathsf{ScaledGeLU}}
\newcommand{\absmodule}{\mathsf{Abs}}
\newcommand{\linear}{\mathsf{Linear}}

\newcommand{\conv}{\mathsf{Conv2D}}
\newcommand{\resmlp}{\mathsf{ResMLP}}
\newcommand{\resnet}{\mathsf{ResNet}}
\newcommand{\gpt}{\mathsf{GPT}}
\newcommand{\embed}{\mathsf{Embed}}
\newcommand{\meansub}{\mathsf{MeanSubtract}}
\newcommand{\RMSdivide}{\mathsf{RMSDivide}}
\newcommand{\tare}{\mathsf{tare}}
\newcommand{\mnew}{m_{\mathrm{new}}}

\newcommand{\Initial}{\mathsf{InputLayer}}
\newcommand{\Block}{\mathsf{Block}}
\newcommand{\Final}{\mathsf{OutputLayer}}

\newcommand{\funcattn}{\mathsf{FuncAttention}}
\newcommand{\shattn}{\mathsf{Attention}}
\newcommand{\mhattn}{\mathsf{MultiHeadAttention}}
\newcommand{\querymod}{\mathsf{Query}}
\newcommand{\keymod}{\mathsf{Key}}
\newcommand{\valuemod}{\mathsf{Value}}
\newcommand{\exitmod}{\mathsf{Exit}}
\newcommand{\mlp}{\mathsf{MLP}}
\newcommand{\RMS}{\mathsf{RMS}}
\newcommand{\inftyRMS}{\infty\mathsf{RMS}}
\newcommand{\inftyop}{\infty\mathsf{-op}}

\newcommand{\rmsdivide}{\mathsf{RMSDivide}}
\newcommand{\attn}{\mathsf{Attn}}

\newcommand{\Loss}{\mathcal{L}}

\newcommand{\modula}{Modula\xspace}
\newcommand{\python}[1]{\texttt{#1}}
\newcommand{\pythoncomment}[1]{\textcolor{Green}{\texttt{#1}}}

\newcommand{\mask}{\bm{\mathrm{mask}}}

\renewcommand{\theequation}{\arabic{section}.\arabic{equation}}
\numberwithin{equation}{section}

\title{Scalable Optimization in the Modular Norm}

\author{%
}

\begin{document}

\maketitle

\vspace{-4.5em}
\begin{tabularx}{\textwidth}{>{\centering\arraybackslash}p{0.28725\textwidth}@{}>{\centering\arraybackslash}p{0.4\textwidth}@{}>{\centering\arraybackslash}p{0.3\textwidth}}
\textbf{Tim Large}$^\star$ & \textbf{Yang Liu} & \textbf{Minyoung Huh} \\
Columbia University & Lawrence Livermore National Lab& MIT CSAIL\\[4ex]
\textbf{Hyojin Bahng}   & \textbf{Phillip Isola} & \textbf{Jeremy Bernstein}$^\star$\footnote[0]{$\star$ denotes equal contribution. Correspondence to \texttt{\{jbernstein,minhuh\}@mit.edu}.} \\
MIT CSAIL & MIT CSAIL & MIT CSAIL\\
\end{tabularx}
\vspace{1em}

\stoptocentries
\begin{abstract}
    To improve performance in contemporary deep learning, one is interested in scaling up the neural network in terms of both the number and the size of the layers. When ramping up the width of a single layer, graceful scaling of training has been linked to the need to normalize the weights and their updates in the ``natural norm'' particular to that layer. In this paper, we significantly generalize this idea by defining the \textit{modular norm}, which is the natural norm on the full weight space of any neural network architecture. The modular norm is defined recursively in tandem with the network architecture itself. We show that the modular norm has several promising applications. On the practical side, the modular norm can be used to normalize the updates of any base optimizer so that the learning rate becomes transferable across width and depth. This means that the user does not need to compute optimizer-specific scale factors in order to scale training. On the theoretical side, we show that for any neural network built from ``well-behaved'' atomic modules, the gradient of the network is Lipschitz-continuous in the modular norm, with the Lipschitz constant admitting a simple recursive formula. This characterization opens the door to porting standard ideas in optimization theory over to deep learning. We have created a Python package called \modula that automatically normalizes weight updates in the modular norm of the architecture. The package is available via \texttt{pip install modula} with source code \href{https://github.com/jxbz/modula}{\texttt{here}}.
\end{abstract}
\section{Introduction}
\label{sec:intro}

Given the practical impact of deep learning systems trained at the largest scale, there is a need for training algorithms that scale gracefully: without instability and---if possible---without manual tuning. However, current best practices for training have developed somewhat organically and do not live on a bedrock of sound numerical analysis. For example, while the Adam optimizer \citep{kingma_adam:_2015} is ubiquitous in the field, errors have been found in its proof of convergence \citep{adam-proof}, and empirically Adam has been found to scale poorly as either the width \citep{yang2021tuning} or the depth \citep{yang2024tensor} of the network is ramped up.

To remedy this situation, a patchwork of learning rate correction factors have recently been proposed \citep{yang2021tuning,yang2024tensor,bordelon2024depthwise,Jelassi2023DepthDO}. The general idea is to retrofit a base optimizer such as Adam or SGD with special correction factors intended to render the optimizer's optimal learning rate invariant to scale. But this situation is not ideal: the correction factors are reportedly difficult to use. \citet{lingle} suggests that this may be due to their ``higher implementation complexity, many variations, or complex theoretical background''. What's more, the correction factors are optimizer-specific, meaning that if one switches to a different optimizer one must either look up or recalculate a separate set of correction factors.

\begin{figure}
    \centering
    \includegraphics[width=\textwidth]{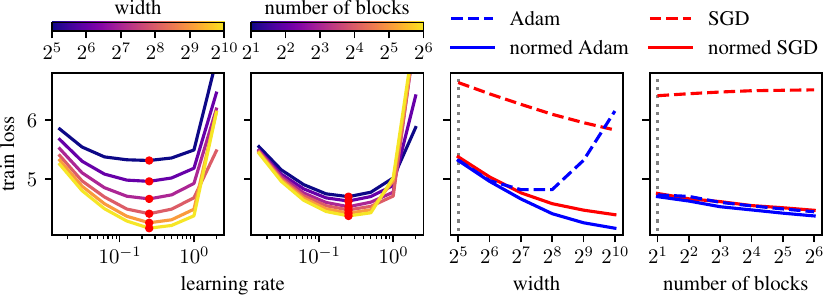}
    \mycaption{Learning rate transfer in the modular norm.}{We train GPT with context length 128 for 10k steps on OpenWebText. \textbf{Left:} Learning rate sweeps for normed Adam (Adam with updates normalized in the modular norm) with three transformer blocks and varying width. The optimal learning rate (marked by red dots) transfers well across scales. \textbf{Mid-left:} The same, but varying the number of blocks at width 128. \textbf{Mid-right:} Comparing normed versus unnormed Adam and SGD at fixed learning rate and varying width. For each method, we tune the learning rate at the scale marked by the dotted line. The normed methods scale better. \textbf{Right:} The same, but scaling number of blocks.}
    \label{fig:openwebtext-transfer}
\end{figure}

The goal of this paper is to simplify matters. We show that both Adam and SGD can be made to scale gracefully with width and depth by simply normalizing their updates in a special norm associated with the network architecture---see \cref{fig:openwebtext-transfer}. We call this norm the \textit{modular norm}, and provide a Python package called \modula that constructs this norm automatically and in tandem with the architecture.

The modular norm is constructed recursively, leveraging the module tree perspective on neural architectures. It is enough to define how the modular norm propagates through only two elementary operations: composition and concatenation. We show how other basic operations on modules, such as addition and scalar-multiplication, can be implemented through composition and concatenation. And then higher-order structures, such as residual networks, can be built using these basic operations.

Beyond its practical relevance, the modular norm may also prove useful to theoreticians. Various optimization-theoretic quantities are accessible and efficiently calculable in the modular norm. For instance, we show that the gradient of any neural network built from ``well-behaved'' atomic modules is Lipschitz-continuous in the modular norm of the architecture. This opens the door to porting several more-or-less textbook optimization theory analyses \citep{fawzi} over to the world of deep learning.

\subsection{Related work}

\paragraph{Metrization} It is by now well-known that deep networks do not easily or naturally admit Lipschitz-continuity or smoothness guarantees in the Euclidean norm \citep{cohen2021gradient,li2023convex,Zhang2020Why,my-fromage}. Researchers have attempted to address this problem: for instance, \citet{my-fromage} propose a distance function called \textit{deep relative trust}, which combines Frobenius norms across network layers. However, deep relative trust is only constructed for the multilayer perceptron and, when used to normalize updates, its employment of the Frobenius norm precludes good width scaling. In contrast, \citet{my-spectral} equip individual layers with the RMS--RMS operator norm, finding this to enable good width scaling. Researchers have also looked at building neural net distance functions outside the context of scalability \citep{juhan,benjamin2018measuring,Neyshabur2015PathSGDPO}.

\paragraph{Asymptotics} The metrization-based approach to scaling developed in this paper contrasts with the tradition of asymptotic scaling analyses---the study of infinite width and depth limits---more common in the deep learning theory literature \citep{yang2021tuning,Yang2021TensorPI,yang2024tensor,bordelon2024depthwise,lee2018deep}. These asymptotic analyses follow an old observation of \citet{radford} that interesting properties of the neural network function space are exactly calculable in the infinite width limit and at initialization. This tradition has continued with asymptotic studies of the neural tangent kernel \citep{NTKjacot} as well as infinite depth limits \citep{li2022the, bordelon2024depthwise, yang2024tensor}. However, there is increasing recognition of the limits of these limits, with researchers now often trying to relax limiting results \citep{lewkowycz2021the,PDLT-2022,Liu2020OnTL}. And ultimately, from a practitioner's perspective, these results can be difficult to make sense of \citep{lingle}. In contrast, our framework eschews any kind of limiting or probabilistic analysis. As a consequence, we believe our framework is simpler, more easily relatable to basic mathematical concepts, and ultimately more relevant to what one may encounter in, say, a PyTorch \citep{pytorch} program.

\paragraph{Majorization} In recent work, \citet{Streeter2022AutomaticallyBT} propose a \textit{universal majorize-minimize algorithm} \citep{Streeter2023UniversalMA}: a method that automatically computes and minimizes a majorizer for any computational graph. Despite its generality, current downsides to the method include its overhead, which can be 2$\times$ per step \citep{google-blog}, as well as the risk that use of a full majorization may be overly pessimistic. Indeed, \citet{namhoon} find that an optimization approach leveraging second-order information converges significantly faster than a majorization-inspired approach. Related ideas appear in \citep{agd-2023,Tran2015FastDT}.
\section{Descent in Normed Spaces}\label[section]{sec:tech_intro}

We define the modular norm in \cref{sec:module_theory}. This section is intended to prime the reader for what is to come. In this section, and the rest of the document, the diamond operator $\diamond$ denotes tensor contraction.

\subsection{What's in a norm?}\label[section]{sec:norm_criteria}

Suppose that we wish to use gradient descent to minimize a loss function $\el:\weights \to \R$ over a weight space $\weights = \R^{N}$. What properties of the loss $\el$ and weight space $\weights$ would we desire for this to be sensible? Three such properties are:
\begin{enumerate}[label=(\roman*)]
    \item the loss function is differentiable, meaning that  the gradient map $\nabla_\vw \el : \weights \to \weights$ exists;
    \item the weight space $\weights$ carries a norm $\smash{\norm{\cdot}: \weights \to \R}$, which need not be the Euclidean norm;
    \item the loss is Lipschitz smooth in the norm $\norm{\cdot}$, with sharpness constant $\lambda >0$, meaning that:
    \begin{equation}\label[inequality]{eq:smooth}
        \el(\vw+\Delta\vw) \leq \el(\vw) + \nabla_\vw \el(\vw)\diamond \Delta \vw + \frac{\lambda}{2} \norm{\Delta \vw}^2.
    \end{equation}
\end{enumerate}
Under these conditions, the weight update given by $\smash{\Delta\vw= \argmin \left[\nabla_\vw \el(\vw)\diamond \Delta \vw + \frac{\lambda}{2} \norm{\Delta\vw}^2\right]}$ is guaranteed to reduce the loss. The particular norm $\norm{\cdot}$ influences the direction of this weight update, while the sharpness constant $\lambda$ influences the size of the update.

In deep learning, we would ideally like the optimal step-size to remain invariant as we scale, say, the width and the depth of the network. Thus, a fundamental problem is to design a norm such that, first, \cref{eq:smooth} actually holds (and is not hopelessly lax), and second, the corresponding sharpness constant $\lambda$ is invariant to the relevant architectural dimensions. If the norm is chosen poorly, the practitioner may end up having to re-tune the step size as the network is scaled up. In this paper, we design a norm for neural networks that meets these requirements: the \textit{modular norm}.

\subsection{Preview of the modular norm}

The weight space of a deep neural network is a Cartesian product $\weights = \weights_1 \times \hdots \times \weights_L$, where $\weights_k$ is the weight space at layer $k$. \citet{my-spectral} consider the problem of metrizing individual layers. For instance, if layer $k$ is a linear layer with weight space $\weights_k=\R^{d_\out \times d_\inn}$, then they equip this layer with the \textit{RMS--RMS operator norm}, $\norm{\cdot}_{\RMS-\RMS}$. This is the matrix norm induced by equipping the input and output space of the layer with the root-mean-square (RMS) vector norm, $\norm{\vx}_{\RMS}^2 \defeq \tfrac{1}{d} \Sigma_i \, \vx_i^2$ for $\vx\in\R^d$. The advantage of this non-standard matrix norm is that it allows one to estimate the amount of feature change induced by a gradient update. In other words, the inequality
\begin{equation}
    \norm{\Delta\mW \vx}_{\RMS} \leq \norm{\Delta\mW}_{\RMS-\RMS} \cdot \norm{\vx}_{\RMS},
\end{equation}
turns out to hold quite tightly when $\Delta \mW$ is a gradient update and $\vx$ is a corresponding layer input. This is because gradient updates to a layer are (sums of) outer products that align with layer inputs.

Once we know how to metrize individual layers, a natural question is: can we combine layer-wise norms to produce a norm on the full weight space $\smash{\weights = \prod_k \weights_k}$ of the network? Na\"ively, there are many ways to do this: one could take any positive linear combination of the layer-wise norms ($L^1$ combination), the square root of any combination of the squared layer-wise norms ($L^2$ combination), and so on. But we want the norm to be useful by the criteria of \cref{sec:norm_criteria}. To this end, we propose the \emph{modular norm} $\norm{\cdot}_\weights$, which ends up as a max ($L^{\infty}$ combination) of scaled layer-wise norms $\norm{\cdot}_{\weights_k}$:
\begin{equation}\label{eq:scaled_max_norm}
    \norm{(\vw_1, \hdots, \vw_L)}_{\weights}
    := \max\left(s_1 \norm{\vw_1}_{\weights_1}, \hdots, s_L \norm{\vw_L}_{\weights_L}\right).
\end{equation}
The positive scalar constants $s_1, \hdots, s_L$ are determined by both the architecture of the network and a set of user-specified ``mass'' parameters. The precise construction of the modular norm, working recursively over the module tree of the network, is given in \cref{sec:module_theory}; there, we also explain how the modular norm satisfies the criteria of \cref{sec:norm_criteria}, and the role played by the mass parameters. For now, let us explain what good the modular norm yields in practice.

\subsection{Normed optimization}
The main practical use of the modular norm is to normalize weight updates. With reference to \cref{eq:scaled_max_norm}, we define the following operation on weight updates $\Delta \vw = (\D \vw_1, \hdots, \D \vw_L) \in \weights$:
\begin{equation}
    \Normalize(\Delta \vw) :=
    \left(
    \frac{\D \vw_1}{s_1 \norm{\D \vw_1}_{\weights_1}},
    \hdots,
    \frac{\D \vw_L}{s_L \norm{\D \vw_L}_{\weights_L}}
    \right).
\end{equation}
Provided none of the $\Delta\vw_k$ are zero, then $\Normalize(\Delta \vw)$ is a unit vector in the modular norm. We propose using $\Normalize$ as a wrapper, along with an explicit learning rate schedule, for any base optimizer such as Adam or SGD. The resulting \emph{normed optimizer} is thus made architecture-aware via the normalize function. In pseudo-code---and actual Modula code---this amounts to:
{\list{}{\leftmargin=0.2in}\item[]
    \python{delta\_w = optim(w.grad())} \hfill \pythoncomment{\# get update from base optimizer}\\
    \python{net.normalize(delta\_w)}    \hfill \pythoncomment{\# normalize update in the modular norm}\\
    \python{w -= eta(step) * delta\_w}         \hfill \pythoncomment{\# apply update with learning rate eta}
\endlist}
We find this wrapper to significantly improve the scalability of the base optimizer. It renders the optimal learning rate roughly invariant to width and depth, with seemingly no cost to accuracy. In some instances, it enables training with a simpler optimizer---for example, training GPT with SGD rather than Adam---thus incurring a smaller memory footprint.

Normalization in the modular norm essentially forces individual layers to learn at specified, regulated rates. We view this as \emph{balancing} learning across the network; no individual layer can learn too fast and destabilize training. This balance is determined by the architecture, along with user-specified mass parameters that provide precise control over the relative learning speed in different submodules.

For a variety of experiments with normed optimization, see \cref{sec:experiments,app:experiments}. But first, we detail the construction of the modular norm along with its core properties.
\section{Constructing the Modular Norm}\label[section]{sec:module_theory}

\begin{figure}
    \centering
    \includegraphics[trim={0 4.1cm 0 0}]{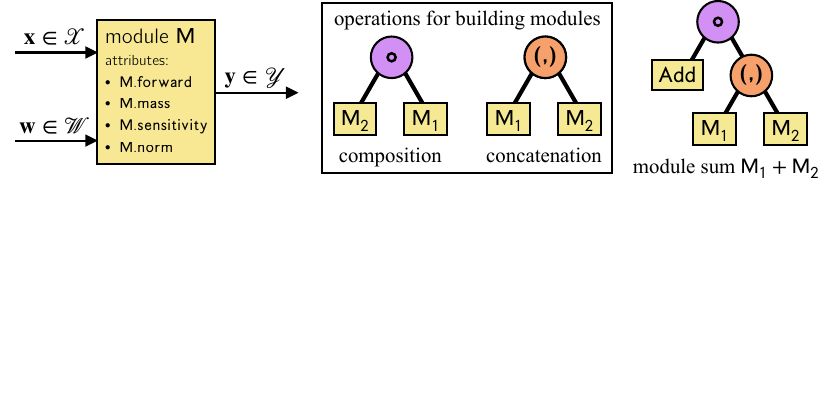}
    \mycaption{Modules and trees of modules.}{A module is an object that maps an input and a weight vector to an output. \textbf{Left:} In addition to the standard \textit{forward} function, our modules are endowed with two numbers---a \textit{mass} and \textit{sensitivity}---and a \textit{norm}. \textbf{Middle:} New \textit{compound modules} are built via the binary operations of composition and concatenation. We provide rules for composing and concatenating all module attributes. \textbf{Right:} Compound modules are binary trees, where the leaves are modules and the internal nodes compose and concatenate their children. Here we illustrate a sum of modules, which leverages a special utility module $\add$---see \cref{tab:operations} for more on this.}
    \label{fig:module}
\end{figure}

Our strategy is to first define the abstract notion of a \textit{module}, which includes a norm as an attribute. We depict this concept in \cref{fig:module}. Then, by providing rules for composing and concatenating modules, we recursively define a norm for any module built via an arbitrary sequence of compositions and concatenations: the modular norm!

\subsection{Modules}

\label{sec:module}
A \textit{module} is a re-usable, composable object useful for building complicated neural networks. Our definition of a module augments the PyTorch module \citep{pytorch} with two real numbers and a norm:
\begin{definition}[Module]\label{def:module} Given input vector space $\inputs$, output vector space $\outputs$ and weight vector space $\weights$, a module $\module$ is an object with the following four attributes:
\begin{enumerate}[label=\normalfont(\alph*)]
\setlength\itemsep{0em}
    \item a function, $\module\for: \weights\times\inputs\to\outputs$, which maps an input and a weight vector to an output---we often abbreviate this attribute to just $\module \equiv \module\for$;
    \item a number, $\module\mass\geq0$, which will turn out to set the proportion of feature learning that this module contributes to any supermodule;
    \item a number, $\module\lip \geq0$, which estimates the module's sensitivity to input perturbations;
    \item a norm over the weight space, $\module\nor : \weights \to \R_{\geq0}$, sometimes abbreviated to just $\norm{\cdot}_\module$.
\end{enumerate}
\end{definition}
Before we say more about the intended roles of these attributes, let us mention the three kinds of modules that we will care about in practice:
\begin{enumerate}[label=(\roman*)]
\item \emph{atomic modules}, whose attributes are hand-declared, and have weights. Examples include linear modules, embedding modules, and convolution modules.
\item \emph{bond modules}, whose attributes are hand-declared, but have no weights. Formally, their weight space is the zero vector space $\weights = 0$. An example is the $\relu$ non-linearity module.
\item \emph{compound modules}, built out of other modules, with automatically inferred attributes.
\end{enumerate}
Note that the space of objects that type-check as a module by \cref{def:module} is vast. Since we need to hand-declare atomic and bond modules in order to build interesting compound modules, we should have an idea of what makes for a ``good'' module. Simply put, a module is good when its attributes are predictive of its behaviour. To formalize this idea, we say that a module is \textit{well-normed} if its forward function, sensitivity, and norm satisfy the following two relationships:

\begin{definition}[Well-normed]\label{def:well-normed}
Let $\module$ be a module on $(\inputs, \outputs, \weights)$, where the input and output spaces have respective norms $\norm{\cdot}_{\inputs}$ and $\norm{\cdot}_{\outputs}$. $\module$ is well-normed if for all inputs $\vx\in\inputs$ and weights $\vw \in\weights$:
\begin{align}
\norm{\nabla_\vw\module\for(\vw,\vx)\diamond\Delta \vw}_\outputs &\leq \module\nor(\Delta \vw) &&\text{for all }\Delta\vw\in\weights;\\
\norm{\nabla_\vx\module\for(\vw,\vx)\diamond\Delta \vx}_\outputs & \leq \module\lip * \norm{\Delta \vx}_\inputs &&\text{for all }\Delta\vx\in\inputs.
\end{align}
\end{definition}
Well-normed-ness means that the norm function and sensitivity are a good match for the forward function. The first inequality says that a well-normed module is Lipschitz-continuous over its weight space with a constant one. The second inequality says that a well-normed module is Lipschitz-continuous over its input space with constant $\module\lip$. In practice, we will be interested in well-normed modules where these inequalities hold fairly tightly, since then $\module\lip$ and $\module\nor$ will let us estimate the sensitivity of the module to input and weight perturbations. \cref{app:module-design} provides many examples of well-normed atomic and bond modules.

The remaining attribute $\module\mass$ will turn out to control the proportion of feature learning that a module contributes to any compound module in which it participates. We formalize this concept in \cref{sec:mass}. But before that, we need to understand how to build compound modules.

\subsection{Compound modules: Building new modules from old}
\label{sec:compound}

We consider building new modules from old ones via the binary operations of composition and concatenation, illustrated in \cref{fig:module}. Composition is denoted via the serial combination $\module_2 \circ \module_1$, and concatenation via the parallel combination $(\module_1, \module_2)$, alternatively referred to as a \textit{module tuple}. These simple binary combinations will let us build basic algebraic operations on modules (\cref{tab:operations}) as well as complex neural network architectures. We start by defining module composition:
\begin{definition}[Module composition]\label{def:composition} Consider module $\module_1$ with input, output and weight space $(\inputs_1,\outputs_1,\weights_1)$ and module $\module_2$ with input, output and weight space $(\inputs_2,\outputs_2,\weights_2)$. $\module_1$ and $\module_2$ are \textit{composable} if $\inputs_2 = \outputs_1$. Their composite $\module=\module_2\circ\module_1$ lives on $(\inputs_1,\outputs_2,\weights_1 \times \weights_2)$ with attributes:
\begin{enumerate}[label=\normalfont(\alph*)]
\setlength\itemsep{0em}
\item $\module\for((\vw_1,\vw_2),\vx)) = \module_2\for(\vw_2,\module_1\for(\vw_1,\vx))$;%

\item $\module\mass = \module_1\mass + \module_2\mass$;%

\item $\module\lip = \module_1\lip * \module_2\lip$;%

\item $\module\nor((\vw_1, \vw_2))$ given by:
\begin{equation*}\label{eq:norm_composition}
     \max\left(
    \module_2\lip * \frac{\module\mass}{\module_1\mass} * \module_1\nor(\vw_1),
    \frac{\module\mass}{\module_2\mass} * \module_2\nor(\vw_2)\right),
\end{equation*}
where if $\module_1\mass$ or $\module_2\mass$ is zero, the corresponding term in the $\max$ is set to zero.
\end{enumerate}
\end{definition}
At this stage, we make two comments about this definition. First, in the definition of the composite norm, notice that the norm of the first module couples with the sensitivity of the second module. This reflects the fact that the output of the first module is fed into the second module and not vice versa. Second, observe that the masses of the submodules are involved in setting the balance of the composite norm. Before we further motivate this definition, let us first define module concatenation:
\begin{definition}[Module concatenation]\label{def:concatenation} Consider module $\module_1$ with input, output and weight space $(\inputs_1,\outputs_1,\weights_1)$ and module $\module_2$ with input, output and weight space $(\inputs_2,\outputs_2,\weights_2)$. We say that $\module_1$ and $\module_2$ are concatenatable if their input spaces match: $\inputs_1 = \inputs_2$. The tuple $\module=(\module_1,\module_2)$ has input, output and weight space $(\inputs_1, \outputs_1\times\outputs_2, \weights_1\times\weights_2)$ and attributes:
\begin{enumerate}[label=\normalfont(\alph*)]
\setlength\itemsep{0em}
\item $\module\for((\vw_1,\vw_2),\vx)) = (\module_1\for(\vw_1,\vx), \module_2\for(\vw_2,\vx))$;%
\item $\module\mass = \module_1\mass + \module_2\mass$;%
\item $\module\lip = \module_1\lip + \module_2\lip$;%
\item $\module\nor(\vw_1, \vw_2)$ given by:
\begin{equation*}\label{eq:norm_addition}
    \max\left(
    \frac{\module\mass}{\module_1\mass} * \module_1\nor(\vw_1),
    \frac{\module\mass}{\module_2\mass} * \module_2\nor(\vw_2)\right),
\end{equation*}
where if $\module_1\mass$ or $\module_2\mass$ is zero, the corresponding term in the $\max$ is set to zero.
\end{enumerate}
\end{definition}

Concatenation is simpler than composition in the sense that neither module is fed through the other, and therefore, sensitivity does not appear in the concatenated norm. To further motivate these definitions, observe that two basic and desirable properties follow as immediate consequences:

\begin{proposition}[Composition and concatenation are associative]\label{prop:assoc} If modules $\module_1, \module_2, \module_3$ are successively composable, then $\module_3 \circ (\module_2 \circ \module_1)$ equals $(\module_3 \circ \module_2) \circ \module_1$ in all attributes. If modules $\module_1, \module_2, \module_3$ are mutually concatenatable, then $((\module_1, \module_2), \module_3)$ equals $(\module_1, (\module_2, \module_3))$ in all attributes.
\end{proposition}

\begin{proposition}[Composition and concatenation preserve well-normedness]\label{prop:wellnormed} If modules $\module_1$ and $\module_2$ are well-normed and composable, then their composite $\module_2\circ\module_1$ is also well-normed. If modules $\module_1$ and $\module_2$ are well-normed and concatenatable, then their tuple $(\module_1,\module_2)$ is also well-normed with respect to the $L^1$ combination norm on the output space: $\norm{(\cdot, \cdot)}_{\outputs_1 \times \outputs_2} = \norm{\cdot}_{\outputs_1} + \norm{\cdot}_{\outputs_2}.$
\end{proposition}
The proofs follow directly from the definitions and the chain rule. \cref{prop:assoc} implies that one may build complicated compound modules without worrying in which order successive combinations are taken. \cref{prop:wellnormed} implies that complicated compounds automatically inherit Lipschitz guarantees. 

Taken together, \cref{def:composition,def:concatenation} define the \textit{modular norm} $\module\nor$ of any compound module $\module$.

\renewcommand{\arraystretch}{1.5}
\begin{table}
    \centering
    \begin{tabular}{|c|c|c|c|c|} \hline
        \textbf{Operation} & \textbf{Shorthand} & \textbf{Definition} & \textbf{\modula Expression}\\ \hline\hline
         
        module addition & $\module_1 + \module_2$ & $\add \circ (\module_1, \module_2)$ & \python{M\_1 + M\_2} \\

        scalar multiplication & $a * \module$ & $\mult_{a} \circ \module$ & \python{a * M}\\

        iterated composition & $\module^L$ & $\module \circ \module^{L-1}$ with $\module^0 \defeq \identity$ & \python{M ** L}\\

         \hline
    \end{tabular}
    \mycaption{Arithmetic with modules.}{Composition and concatenation let us define an extended arithmetic on modules. The utility modules $\add, \mult_a$ and $\identity$ are defined in \cref{sec:bonds}.}
    \label{tab:operations}
\end{table}

\subsection{Mass allocation in compound modules}
\label{sec:mass}

Suppose we wish to train a network with an input layer, an output layer, and $L$ blocks between:
\begin{align}
    \mathsf{Network} &= \mathsf{OutputLayer} \circ \mathsf{HiddenLayers} \circ \mathsf{InputLayer}\\ &= \mathsf{OutputLayer} \circ \mathsf{Block}^L \circ \mathsf{InputLayer}.
\end{align}
Then how much learning should happen in the output layer, compared to the blocks, compared to the input layer? And what if we scale the number of blocks $L$---do we want relatively less learning to occur in the network's extremities? Or do we want the input and output layers to learn non-trivially even in the $L\to\infty$ limit? Since answering these questions is difficult a priori, we introduced the mass parameter to allow a user to set the proportional contribution each module has toward learning:
\begin{proposition}[Feature learning is apportioned by mass]\label{prop:character}
    Consider a compound module $\module$ derived in any fashion from $L$ well-normed modules $\module_1, \hdots, \module_L$. Given weight setting $\vw = (\vw_1, \hdots, \vw_L)$, where $\vw_k$ denote the weights of module $\module_k$, let us perturb $\vw$ by $\D \vw = (\D \vw_1, \hdots, \D \vw_L)$. If we decompose the linearized change in the output of module $\module$ into one contribution per sub-module:
\begin{equation}
    \nabla_\vw \module(\vw,\vx) \diamond \Delta \vw = \nabla_{\vw_1} \module(\vw,\vx) \diamond \Delta \vw_1 + \dots + \nabla_{\vw_L} \module(\vw,\vx) \diamond \Delta \vw_L,
\end{equation}
then the $k$th term in this decomposition satisfies:
\begin{equation}
    \norm{\nabla_{\vw_k} \module(\vw,\vx) \diamond \D \vw_k}_{\outputs} \le \frac{\module_k \mass}{\module\mass} * \module\nor(\D \vw).
\end{equation}
\end{proposition}

In words: module mass provides the flexibility needed to build complicated compound modules involving many sub-modules, while maintaining precise control over how much learning any sub-module can contribute to the overall compound. \cref{prop:character} is proved in \cref{sec:massproof}.

\begin{figure}
    \begin{subfigure}[t!]{.64\linewidth}
      \centering
      \includegraphics[width=\linewidth]{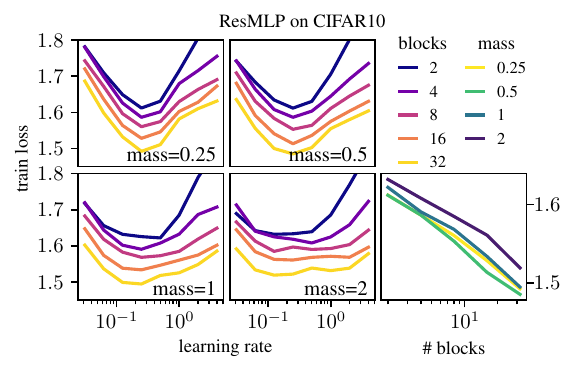}
    
      \label{fig:sub1}
    \end{subfigure}%
    \hfill
    \begin{subfigure}[t!]{.353\linewidth}
      \centering
      \includegraphics[width=1.0\linewidth]{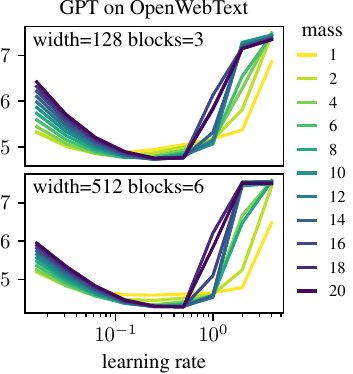}
      
      \label{fig:sub2}
    \end{subfigure}
    \mycaption{Exploring mass allocation.}{We tune the total mass of the hidden layers, training with normed Adam. \textbf{Left group:} Learning rate sweeps for ResMLP on CIFAR-10, for varying depth and mass. The bottom right subplot reports the best train loss at each mass and depth. Mass 0.5 is best at all depths. \textbf{Right group:} Learning rate sweeps for GPT on OpenWebText, for varying mass. Both optimal mass and learning rate transferred from the small model (top) to the large model (bottom).}
    \label{fig:mass}
\end{figure}

In practice, we obtained the best training performance by maintaining a constant amount of learning in the input and output layers even as the number of blocks is scaled (\cref{fig:mass-sweep-resmlp-appendix}). In other words, it seems to be a good idea to assign $\mathsf{OutputLayer}\mass : \mathsf{HiddenLayers}\mass : \mathsf{InputLayer}\mass $ in proportion $1 : m : 1$, where $m$ is independent of the number of blocks $L$. The exact mass of the hidden layers $m$ needs to be tuned on a new architecture---just as one needs to tune separate learning rates in the input and output layers in $\mu$P \citep{Yang2021TensorPI}; this tuning can be done on a small model prior to scaling (\cref{fig:mass}). We further discuss mass allocation in \cref{app:mass_allocation}.

\subsection{Smoothness in the modular norm}
\label{sec:smoothness}

In this section, we study the second derivatives of a module using the modular norm as a measuring stick. Let us start by defining the notion of sharpness that we will consider:

\begin{definition}[Module sharpness]\label{def:sharp} Let $\module$ be a module on $(\inputs, \outputs, \weights)$, where the input and output spaces have respective norms $\norm{\cdot}_{\inputs}$ and $\norm{\cdot}_{\outputs}$. We say that $\module$ is $(\alpha, \beta, \gamma)$-sharp for constants $\alpha, \beta, \gamma \geq 0$ if, at all inputs $\vx \in \inputs$ and weights $\vw \in \weights$, the second derivatives of $\module$ are bounded as:
\begin{align}
    {%
    }
    \norm{\Delta \vw \diamond \nabla^2_{\vw\vw}\module(\vw,\vx)\diamond\Delta \widetilde{\vw}}_{\outputs} &\leq \alpha \,\norm{\Delta \vw}_{\module} \norm{\Delta \widetilde{\vw}}_{\module} && \text{for all }\Delta\vw, \Delta \widetilde{\vw}\in\weights;\\
    \norm{\Delta \vw \diamond \nabla^2_{\vw \vx}\module(\vw,\vx)\diamond\Delta \vx}_{\outputs} &\leq \beta\,\norm{\Delta \vw}_{\module}\,\norm{\Delta \vx}_{\inputs} &&\text{for all }\Delta\vw\in\weights \text{ and } \Delta\vx\in\inputs;\\
    {%
    }
    \norm{\Delta \vx \diamond \nabla^2_{\vx\vx}\module(\vw,\vx)\diamond\Delta \widetilde{\vx}}_{\outputs} &\leq \gamma\, \norm{\Delta \vx}_{\inputs}\norm{\Delta \widetilde{\vx}}_{\inputs} &&\text{for all }\Delta\vx, \Delta \widetilde{\vx} \in\inputs.
\end{align}
\end{definition}
While one may ultimately be interested in the sharpness of a module with respect to weight perturbations, \cref{def:sharp} also tracks sharpness with respect to input perturbations. In fact, tracking this extra information is essential for propagating sharpness bounds up the module tree. \cref{app:sharpness} details the procedure for automatically calculating the sharpness constants of a compound module starting from the sharpness constants of all its submodules; see \cref{prop:sharp_comp,prop:sharp_concat} for the specific formulae. Here we highlight one major corollary of these formulae, proved in \cref{app:res-proof}: \emph{for a specific choice of block multipliers, the sharpness constant of a residual network is independent of depth}:
\begin{proposition}\label{thm:sharpness_residual_modules}Suppose $\module$ is a well-normed, $(\alpha, \beta, \gamma)$-sharp module on $(\inputs, \inputs, \weights)$ with unit sensitivity. Define the depth $L$ \emph{residual module} $\residual{L}{\module}$ via the module arithmetic of \cref{tab:operations} as:
\begin{equation}\label{eq:residual}
    \residual{L}{\module} := \left(\tfrac{L-1}{L} * \identity + \tfrac{1}{L} * \module\right)^L.
\end{equation}
Then this residual module $\residual{L}{\module}$ is in fact $(\alpha + \beta + \tfrac{\gamma}{3}, \beta + \tfrac{\gamma}{2}, \gamma)$-sharp, independent of depth $L$.
\end{proposition}

For optimization purposes, one may be more interested in the sharpness of the loss function rather than the sharpness of the neural network. Fortunately, it is possible to convert sharpness bounds on modules into sharpness bounds on loss functions, provided a little is known about the error measure:

\begin{proposition}[Loss functions are smooth in the modular norm]\label{thm:smoothness} Let $\module$ be a module on $(\inputs, \outputs, \weights)$ and let $\ell : \outputs \times \targets \to \R$ measure the error between a module output and a target in target space $\targets$. The loss $\Loss : \weights \to \R$ records the module's average error on data distribution $\data$ over $\inputs \times \targets$:
\begin{equation}\Loss(\vw) \defeq \Expect_{\vx,\vt \sim \data} \ell(\module(\vw, \vx), \vt).\end{equation}
Suppose that the error measure $\ell$ is $\sigma$-Lipschitz and $\tau$-smooth in the module output, in the sense that:
\begin{align}\abs{\nabla_{\vy} \ell(\vy, \vt) \diamond \D \vy} &\leq \sigma \, \norm{\D \vy}_{\outputs} &&\textnormal{for all } \D \vy \in \outputs \textnormal{ and } \vt \in \targets;\\
\abs{\D \vy \diamond \nabla^2_{\vy \vy} \ell(\vy,\vt) \diamond \D \widetilde{\vy}} &\leq \tau \, \norm{\D \vy}_{\outputs}\,\norm{\D \widetilde{\vy}}_{\outputs} &&\textnormal{for all } \D \vy, \D \widetilde{\vy} \in \outputs \textnormal{ and } \vt \in \targets.
\end{align}
If the module $\module$ is well-normed and $(\alpha,\beta,\gamma)$-sharp, then the loss function $\el$ satisfies the following three inequalities at all weight settings $\vw \in\weights$ and for all weight perturbations $\Delta \vw, \D \widetilde{\vw} \in \weights$:
\begin{enumerate}[label=\normalfont(\roman*)]
    \item $\abs{\D \vw \diamond \nabla^2_{\vw \vw} \Loss \diamond \D \widetilde{\vw}}
    \leq (\sigma \alpha + \tau)\,\norm{\D \vw}_{\module}\,\norm{\D \widetilde{\vw}}_{\module};$
    \item $\norm{\nabla_{\vw} \Loss(\vw +\Delta\vw) - \nabla_{\vw} \Loss(\vw)}_\module^* \leq (\sigma \alpha + \tau)\, \norm{\Delta \vw}_\module$,
    
    where $\norm{\cdot}_\module^*$ is the dual norm of $\norm{\cdot}_\module$;
    \item $\left|\el(\vw+\Delta\vw) -\left[\el(\vw) + \nabla_\vw \el \diamond \Delta \vw \right]\right|\leq  \tfrac{1}{2}(\sigma \alpha + \tau)\,\norm{\D \vw}_{\module}^2 .$
\end{enumerate}
\end{proposition}

The proof is given in \cref{app:smoothness}, and we present estimates for $\sigma$ and $\tau$ for common error measures in \cref{app:errors}. Notice that inequalities (i), (ii) and (iii) are the standard inequalities of smooth optimization \citep{fawzi}, albeit expressed in the modular norm. In fact, (i) implies (ii) implies (iii). In words, inequality (ii) says that the gradient of the loss is Lipschitz-continuous in the modular norm. The Lipschitz constant depends on the module only through the module's first sharpness coefficient $\alpha$.
\section{Experiments}\label{sec:experiments}

Our experiments aimed to test the \emph{scalability of training with normed versions of Adam and SGD}: whether one can tune the learning rate on a small model, and expect the learning rate to remain close to optimal on models of much larger width and depth. In addition to the learning rate, normed optimization in Modula requires a \emph{mass parameter} to apportion feature learning between the input, output and hidden layers; we also tested the sensitivity of this parameter, whether it affects learning rate transfer, and to what extent the optimal mass itself transfers across width and depth.

All SGD experiments were done with momentum $\beta = 0.9$, and all Adam experiments used $\beta_1 = 0.9$ and $\beta_2 = 0.99$. No weight decay was used in any experiment. Every experiment was done with a linear decay learning rate schedule. As for initialization, we used orthogonal initialization for $\linear$ and $\conv$ modules, and Gaussian weights projected to a unit norm ball for our $\embed$ module. This was to ensure all modules were well-normed at initialization. Precise versions of our architectures are described in Appendices \ref{app:network-design} and \ref{app:transformer}. We compare with nanoGPT using standard initialization in \cref{app:std_comparison} to make sure our changes recover standard performance. We actually found unnormed Adam using our GPT architecture transferred learning rate \textit{better} than in nanoGPT.

\begin{figure}
    \centering
    \includegraphics[width=\textwidth]{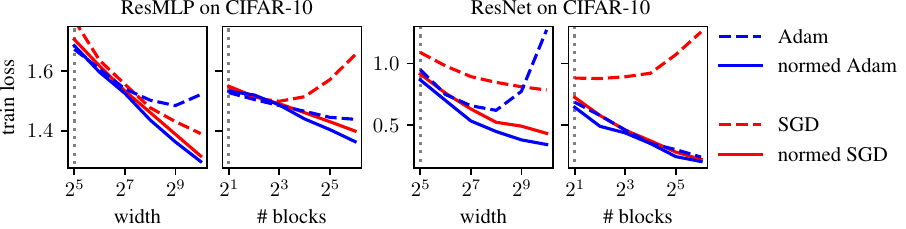}
    \mycaption{Learning rate transfer on CIFAR-10.}{We tune the learning rate on a small model---at the scale marked by the dotted line---and test the performance on models of increasing width and depth at this fixed learning rate. We find that normed Adam and SGD scale better than their unnormed counterparts on both ResMLPs and ResNets. See \cref{fig:openwebtext-transfer} for the same experiment on GPT.}
    \label{fig:vision_results}
\end{figure}

We found that normed optimization, with both Adam and SGD as the base optimizer, allows for successful learning rate transfer across width and depth for GPT training on OpenWebText (\cref{fig:openwebtext-transfer}), as well as ResMLP and ResNet training on CIFAR-10 (\cref{fig:vision_results}). We present expanded results in \cref{app:full-sweep}, including results on test loss. We reproduce the standard finding that train and test loss are remarkably simillar in large language model pretraining. As for mass allocation, \cref{fig:mass} shows that optimal mass transfers with depth for training a ResMLP on CIFAR-10 with normed Adam, and also that both mass and learning rate transfer quite well from a smaller GPT on OpenWebText to a larger one. We detail more experiments on mass allocation in \cref{app:mass_allocation}.

\section{Discussion: Limitations and Future Work}
\label{sec:discuss}

This paper was influenced by four main streams of work: first, the Tensor Programs series, starting at TP-IV \citep{yang2021tuning,Yang2021TensorPI,yang2024tensor}; second, the papers on universal majorize-minimize algorithms \citep{Streeter2022AutomaticallyBT,Streeter2023UniversalMA}; third, work on deep network metrization \citep{my-fromage,my-spectral,agd-2023}; and fourth, the open source deep learning ecosystem \citep{pytorch,jax2018github,Theano} including the PyTorch module tree and Karpathy's YouTube video on autograd  \citep{karpathy-micrograd}. We have distilled and synthesized key ideas from these sources, creating a framework that we believe to be simpler than Tensor Programs, computationally lighter than universal majorization-minimization, more general than prior work on metrization and more scalable than the PyTorch module tree. We have packaged these ideas into a (soon-to-be) open-source library called Modula. Inevitably, Modula has limitations. We highlight some of them here, along with associated avenues for future work.

\textbf{Loss of well-normed-ness.} We have emphasized well-normed-ness (\cref{def:well-normed}) as an important criterion in module design. We show in \cref{sec:atoms} that, for example, the $\linear$ module is well-normed when its weights lie within a spectral norm ball. In our experiments, we initialize all weights so that all modules are well-normed, but we do not enforce this property throughout training. Future work could explore regularization as a means to enforce well-normed-ness throughout training, with the hope of attaining better scalability or improved generalization.

\textbf{Overhead of normalization.} As discussed in \cref{sec:power}, we implement normalization for $\linear$ and $\conv$ modules using two steps of online power iteration. While online power iteration is an established and fast primitive in deep learning---in fact, coming from the GAN literature \citep{miyato2018spectral}---it does add a modest overhead to training time, as discussed in \cref{sec:overhead}. We think it may be possible to mitigate this overhead by constructing atomic modules with more exotic operator norms. For example, if one equips feature vectors with the $L^\infty$ norm rather than the RMS norm, then the induced $L^\infty$--$L^\infty$ matrix norm is cheaper to compute than the RMS--RMS operator norm. In fact, $L^\infty$--$L^\infty$ operator normalization has the convenient feature that it decouples over matrix rows, making it more \textit{local} than spectral normalization and, dare-we-say, more \textit{biologically plausible}.

\textbf{Automatic step-size selection.} Beyond scalability, recent work has explored the question of automatic learning rate selection \cite{mishchenko2024prodigy,pmlr-v202-defazio23a,Ivgi2023DoGIS,agd-2023}, with the Prodigy optimizer \citep{mishchenko2024prodigy} serving as a popular example. We tested the Adam version of Prodigy and found it performs well at small scales, essentially working by an implicit form of line search. However, Prodigy will always break at large enough widths, since it requires a lower bound ($d_0$) on Adam's initial learning rate; \citet{yang2021tuning} showed that no such lower bound exists. We believe this issue could be fixed by rebuilding Prodigy on top of Modula. More broadly, we think that designing line search methods in a properly-normed space is a good idea.

\section*{Acknowledgements}
We are grateful to Chris Mingard, Virgile Richard and Evan Kiely for useful discussions early in the project. Tongzhou Wang and Jyo Pari provided helpful feedback on the writing and figures.

The work was supported by a Packard Fellowship and a Sloan Research Fellowship to PI, by the MIT-IBM Watson AI Lab, by ONR MURI grant N00014-22-1-2740 and the MIT Quest for Intelligence. TL was supported by a Simons Junior Fellowship.

\section*{Contribution Statement}
All authors were involved in project conception and discussions, which were initiated by JB. TL developed the theory with input from JB. MH and YL made core experimental observations. YL, MH, JB, and HB ran experiments. TL and JB did most of the writing, while JB, MH and YL made the figures. PI contributed guidance and helpful feedback throughout the course of the project. JB wrote the Modula package with help from MH.

\clearpage
\bibliographystyle{style/unsrtnat}
\bibliography{refs}

\clearpage

\appendix
\renewcommand{\thesection}{Appendix \Alph{section}}
\renewcommand{\thesubsection}{\Alph{section}.\arabic{subsection}}
\renewcommand{\theequation}{\Alph{section}.\arabic{equation}}

\renewcommand{\contentsname}{Contents of the Appendices}
\tableofcontents
\addtocontents{toc}{\vspace{0pt}}
\clearpage

\starttocentries
\section{The Modula Package}
\label[appendix]{app:modula}

We created a Python package called \modula that realizes our module framework in code. Modula supplements PyTorch's \python{Tensor} class with two new classes: \python{Vector} and \python{Module}.

\subsection{The \python{Vector} class}

The \python{Vector} class is used to store the weights of a module. It allows for basic algebraic operations to be performed on module weights without needing to write \python{for} loops over lists of tensors. For example, if \python{v\_1} and \python{v\_2} are vectors with the same sub-structure, then one may write expressions such as \python{v\_1 + v\_2} for the vector sum, or \python{v\_1 * v\_2} for the elementwise product. Internally, a \python{Vector} stores a list of tensors and implements operations using efficient PyTorch \python{foreach} primitives.

\subsection{The \python{Module} class}

The most significant aspect of the \modula package is the \python{Module} class. A \python{Module} must have six attributes: two \python{float} attributes, namely \python{mass} and \python{sensitivity}. And four methods:
\begin{itemize}[leftmargin=2em]
    \item \python{forward(w: Vector, x: Tensor) -> Tensor} \hfill\pythoncomment{\# returns an output tensor}
    \item \python{initialize() -> Vector} \hfill\pythoncomment{\# randomly samples a weight vector}
    \item \python{normalize(w: Vector)} \hfill\pythoncomment{\# normalizes w to have unit modular norm}
    \item \python{regularize(w: Vector, strength: float)} \hfill\pythoncomment{\# regularizes w in-place}
\end{itemize}
The norm of a module is not specifically implemented, instead we use the normalize method which is how the norm is directly used in optimization.

We refer to modules with hand-specified attributes as \textit{bonds} if they have no weights and \textit{atoms} if they have weights. Modules formed by combining existing modules are called \textit{compounds}. Modula automatically constructs the attributes of compound modules. We provide reference implementations for many common modules---see \cref{app:module-design}. We equip atoms with their natural operator norm, and compute spectral norms via online power iteration. Reference modules may be imported as follows:

{\list{}{\leftmargin=0.2in}\item[]
    \python{from modula.bond~~~~~~import Identity, ReLU, Abs, FunctionalAttention}\\
    \python{from modula.atom~~~~~~import Linear, Embed, Conv2D}\\
\python{from modula.compound~~import ResMLP, ResCNN, Attention, GPT}
\endlist}

To make building new compounds easier, Modula overloads the following operations on modules:
\begin{itemize}[leftmargin=2em]
    \item \python{M\_2 @ M\_1} \hfill \pythoncomment{\# composes module M\_2 with module M\_1}
    \item \python{(M\_1, M\_2)} \hfill \pythoncomment{\# acts as a tuple module in any further composition}
    \item \python{M\_1 + M\_2} \hfill \pythoncomment{\# returns the module sum}
    \item \python{a * M} \hfill \pythoncomment{\# multiplies module M by scalar a}
    \item \python{M ** L} \hfill \pythoncomment{\# returns the Lth iterate of module M}
\end{itemize}
For example, the compound
\begin{equation*}
    \text{\python{(L/(L-1) * Identity() + 1/L * M()) ** L}}
\end{equation*}
builds an \python{L}-layer residual network from base module \python{M}. Comparing with \cref{eq:residual}, we see that \modula expressions closely resemble their mathematical counterparts.

Finally, all modules come with a convenience method \python{tare(m: float)}, which resets the module mass to \python{m}, with default \python{m=1}.

\subsection{Normalization in Modula}
\label[subappendix]{sec:power}

We can normalize any base optimizer in the modular norm using the following pattern:

{\list{}{\leftmargin=0.2in}\item[]
    \python{delta\_w = optim(w.grad())} \hfill \pythoncomment{\# get update from base optimizer}\\
    \python{net.normalize(delta\_w)}    \hfill \pythoncomment{\# normalize update in the modular norm}\\
    \python{w -= lr * delta\_w}         \hfill \pythoncomment{\# apply update to weights}
\endlist}

Computation of \python{net.normalize(delta\_w)} requires an efficient estimation of the spectral matrix norm, in the last two dimensions, of the constituent tensors of \python{delta\_w}; this can be done very quickly to reasonable accuracy using power iteration. We implement this by storing a running estimate of the top singular vector \python{u} for each constituent tensor of \python{delta\_w}. At initialization, \python{u} is sampled Gaussian, and each time we normalize a weight update, the previous update's estimated singular vector is used as the starting value for the power iteration. This enables us to use just two steps of power iteration per weight update. Indeed, for any base optimizer with momentum, successive weight updates should be fairly close; for training without momentum more steps of power iteration may be required.

\subsection{Overhead}
\label[subappendix]{sec:overhead}

To test the overhead of normalization in the modular norm, we trained a width 64 ResMLP with 8 blocks and block-depth 2 for 10k steps on the CIFAR-10 dataset. We repeated the experiment with and without normalization, and in each case with three different random seeds. Without normalization, the training took $101 \pm 1$ seconds, and with normalization the training took $124 \pm 1$ seconds. So in this experiment, the overhead of modular normalization was around 23\%.

We note that the user of the Modula package is free to write new atomic modules with cheaper or more efficient normalize functions. For instance, the Frobenius norm can be used as a proxy for the spectral norm whenever the weight updates have low stable rank \citep{agd-2023,my-spectral}. And we note in \cref{sec:discuss} that one could explore more exotic norms such as the $L^\infty$--$L^\infty$ operator norm, which is cheaper to compute than the standard spectral norm. Beyond these suggestions, one could explore CUDA-level optimizations to spectral norm computation, which is something that we have not explored.
\clearpage
\section{Module and Network Design}
\label[appendix]{app:module-design}

In this appendix, we list the basic, hand-declared modules that serve as building blocks for more complicated neural networks. Then we go on to show how these modules may be combined to yield interesting neural networks. This includes discussion of module broadcasting (\cref{app:broadcasting}) and mass taring (\cref{app:taring}). The appendix culminates with case studies on attention (\cref{app:attention}) and transformers (\cref{app:transformer}).

\subsection{Atomic modules}
\label[subappendix]{sec:atoms}
An \emph{atomic module} or \emph{atom} for short is a module with \emph{nonzero mass and nonzero parameter space}, whose attributes are specifically declared rather than derived. Setting an atom's mass to zero has the effect of freezing its weights under normed optimization.

\renewcommand{\arraystretch}{2}
\begin{table}
    \centering
    \begin{tabular}{|c|c|c|c|c|} \hline
        \textbf{Module} $\bm\module$ & $\bm{\module\for}$ &$\bm{\module\mass}$ & $\bm{\module\lip}$&$\bm{\module\nor}$ \\[0.2ex] \hline \hline
         
        $\linear$ & $\mW,\vx\mapsto \sqrt{\frac{d_\out}{d_\inn}}\,\mW\vx$ &1 & 1 & $\mW \mapsto \norm{\mW}_*$\\
         
         $\embed$ & $\mE,\vx\mapsto\sqrt{d}\,\mE \vx$& 1 & 1 & $\mE\mapsto\max_i \norm{\mE_{\cdot i}}_2$\\
                  
         $\conv$ &$\mC,\vx \mapsto \frac{1}{K^2}\sqrt{\frac{d_\out}{d_\inn}}\,\mC \circledast \vx$&1&1& $\mC \mapsto\max_{ij
         }\norm{\mC_{\cdot\cdot ij}}_*$\\[0.2ex]
         \hline
    \end{tabular}
    \mycaption{Three atomic modules.}{These are the three atoms implemented in \python{Modula}---enough to build ResNet and GPT networks. By including explicit dimensional scale factors in the forward functions, we are able to use the standard spectral norm $\norm{\cdot}_*$ and Euclidean norm $\norm{\cdot}_2$, rather than their rescaled versions. $d_\inn$ and $d_\out$ denote the input and output dimension of the $\linear$ module. $d$ denotes the embedding dimension of the $\embed$ module. $K$ denotes the kernel size of a $\conv$ module with $d_\out$ output channels and $d_\inn$ input channels. $\circledast$ denotes convolution.}
    \label{tab:atoms}
\end{table}

\begin{atomic}[$\linear$]\label{mod:linear}
For positive integers $d_{\out}$ and $d_{\inn}$, the \emph{linear module} $\linear(d_{\out}, d_{\inn})$ corresponds to the standard linear layer with $d_{\inn}$ input features and $d_{\out}$ output features. As a module, it has input space $\inputs = \R^{d_{\inn}}$, output space $\R^{d_{\out}}$ and weights $\weights = \R^{d_{\out} \times d_{\inn}}$ the space of $d_{\out} \times d_{\inn}$ matrices.

Its four attributes (forward function, mass, sensitivity, norm) are given in \cref{tab:atoms}. Note the presence of the $\sqrt{d_{\out}/d_{\inn}}$ factor in the forward function: this convention means that we can work with the \emph{standard} $L^2$ operator norm $\norm{\cdot}_*$ rather than the RMS-RMS operator norm.

Writing $\vf = \linear(d_{\out}, d_{\inn})\for$, its derivative and second derivative at $(\mW, \vx)$ are given by:
\begin{align}
    \nabla \vf \diamond (\D \mW, \D \vx) &= \sqrt{d_{\out}/d_{\inn}}\left((\D \mW) \, \vx + \mW \, (\D \vx)\right), \\
    (\D \mW, \D \vx) \diamond \nabla^2 \vf \diamond (\D \widetilde{\mW}, \D \widetilde{\vx})
    &= \sqrt{d_{\out}/d_{\inn}} \left((\D \mW)(\D \widetilde{\vx}) + (\D \widetilde{\mW})(\D \vx)\right). 
\end{align}
from which we conclude that $\linear(d_{\out}, d_{\inn})$ is well-normed, using the RMS norms on its input and output, so long as its arguments satisfy:
    \begin{equation}\norm{\mW}_{*}, \norm{\vx}_{\RMS} \le 1.\end{equation}
These conditions will be automatically satisfied for many neural networks under \emph{orthogonal initialization} of the weights, and especially if a linear module is immediately preceded by something like a $\layernorm$ module. Moreover, orthogonal initialization guarantees that the well-normed inequality
\begin{equation}
    \norm{\nabla \vf \diamond \D \vx}_{\RMS} \le \norm{\vx}_{\RMS}
\end{equation}
holds tightly in nearly-square matrices at initialization, which is important for getting good signal propagation through the whole network.

Moreover, inspection of the second derivative formula above shows it is always $(0,1,0)$-sharp with respect to the RMS norms on the input and output spaces.\\
\end{atomic}

\begin{atomic}[$\embed$]\label{mod:embedding}
For positive integers $n$ and $d$, the \emph{embedding module} $\embed(n,d)$ corresponds to $n$ class, token, or positional embeddings in a $d$-dimensional embedding space. As a module, it has input space $\R^n$, output space $\R^d$ and weights $\weights = \R^{d \times n}$ the space of $d \times n$ matrices. 

Its attributes are listed in \cref{tab:atoms}.

This is at first sight similar to the linear module, the key difference being that in applications \emph{we expect the inputs of $\embed(n,d)$ to be one-hot vectors}; as such we consider its input space to carry the $L^1$-norm.

As for the linear module, $\embed(n,d)$ is well-normed and $(0,1,0)$-sharp with respect to the $L^1$-norm on the input space $\R^n$ and the RMS norm on the output space $\R^d$.\\
\end{atomic}

\begin{atomic}[$\conv$]\label{mod:conv2d}For positive integers $d_{\out}, d_{\inn}, K$ as well as $H, W$, the \emph{2D-convolution module} $\conv(d_{\out}, d_{\inn}, K)$ corresponds to a convolutional layer with a $K \times K$ kernel; $d_{\inn}$ and $d_{\out}$ are the number of channels for the input and output respectively (we suppress optional stride and padding arguments here for simplicity). Its input space is $\inputs = \R^{d_{\inn} \times H \times W}$, its output space is $\outputs = \R^{d_{\out} \times H \times W}$ and its weights are $\weights = \R^{d_{\out} \times d_{\inn} \times K \times K}$.

Its attributes are listed in \cref{tab:atoms}.

In fact, one could alternatively build $\conv(d_{\out}, d_{\inn}, K)$ starting from $K^2$ different $\linear(d_{\out}, d_{\inn})$ modules (of mass $1/K^2$ each) and concatenating them, and composing with a (parameter-less) convolution module. As such, $\conv$ is well-normed and $(0,1,0)$-sharp. However, in our Modula package, we choose to explicitly declare $\conv$ so as to take advantage of Pytorch's efficient implementation of convolution; the presentation here reflects this.
\end{atomic}

\subsection{Bond modules}
\label[subappendix]{sec:bonds}

A \emph{bond module} or \emph{bond} is a module with zero mass and zero parameter space. They are the ``glue'' between the atomic modules, needed to construct complex neural networks.

Note that \emph{we need not specify a weight space, or mass or norm arguments} for a bond module. Moreover, when discussing whether a bond module is $(\alpha,\beta,\gamma)$-sharp, the inequalities for $\alpha$ and $\beta$ are vacuous; thus for bond modules we will abbreviate this notion to $\gamma$-sharp.

To begin, we need two bond modules that are essentially ``utility'', as they are crucial for defining basic secondary module operations. These modules are also ``type polymorphic'' in the sense that they work with any underlying vector space.

\begin{bond}[$\add$]\label{mod:add}
For any vector space $\outputs$, the \emph{adder module} $\add$ has inputs $\outputs \times \outputs$ and outputs $\outputs$. It has forward function
\begin{equation}\add\for : (\vy_1, \vy_2) \mapsto \vy_1 + \vy_2\end{equation}
and sensitivity 1. Its significance is that it allows for \emph{concatenable modules to be added}:
\begin{equation}
    \module_1 + \module_2 := \add \circ (\module_1, \module_2).
\end{equation}
For any norm $\norm{\cdot}_\outputs$ on the vector space $\outputs$, $\add$ is well-normed with respect to the $L^1$ combination norm $\norm{(\vy_1, \vy_2)}_{\outputs \times \outputs} := \norm{\vy_1}_{\outputs} + \norm{\vy_2}_{\outputs}$ on its input space. Furthermore, $\add$ is $0$-sharp.
\end{bond}

\begin{bond}[$\mult_\lambda$]\label{mod:mult}
For any normed vector space $\outputs$ and real number $\lambda$ the \emph{scalar multiplier module} $\mult_{\lambda}$ has inputs $\outputs$ and outputs $\outputs$. Its forward function is:
\begin{equation}\mult_\lambda\for : \vy \mapsto \lambda * \vy\end{equation}
and its sensitivity is $\abs{\lambda}$. Its significance is that it allows for \emph{scalar multiplication of modules}:
\begin{equation}
    \lambda * \module := \mult_{\lambda} \circ \module.
\end{equation}
It is well-normed with respect to any norm on $\outputs$, and $0$-sharp. When $\lambda = 1$, we call this the \emph{identity module} $\identity = \mult_1$. Note that $\lambda * \identity = \mult_{\lambda}$ for any $\lambda$.
\end{bond}

The remaining bond modules are used explicitly as non-linearities in neural networks.

\begin{bond}[$\absmodule$]
In any dimension $d$, the absolute value bond module $\absmodule$ has inputs and outputs $\R^d$, forward function
\begin{equation}
    \absmodule\for : (x_1, \hdots, x_d) \mapsto (\abs{x_1}, \hdots, \abs{x_d})
\end{equation}
and sensitivity 1. It is well-normed for any norm on $\R^d$.
\end{bond}

\begin{bond}[$\relu$ and $\scaledrelu$]
In any dimension $d$, we define the ``rectified linear unit'' bond module $\relu$ to have input space $\inputs \subset \R^d$, output space $\outputs = \R^d$, forward function
\begin{equation}
    \relu\for : (x_1, \hdots, x_d) \mapsto (\max(0, x_i))_{i=1,\hdots,d}.
\end{equation}
and sensitivity $1/\sqrt{2}$. For this choice of sensitivity, $\relu$ is not well-normed with input space $\inputs$ set to the full $\R^d$. However, it is well-normed if the input space is, informally, a set of dense vectors with balanced signs. For illustration, $\relu$ is rigorously well-normed with respect to the input space 
\begin{equation}
    \inputs = \{\sign{\vt} : \text{for }\vt \in \R^d \text{ with }\# \text{ positive entries }= \# \text{ negative entries}\},
\end{equation}
and RMS norm on inputs and ouputs. For more on this design decision, see \citep{He2015DelvingDI}. We also define $\scaledrelu \defeq \sqrt{2} * \relu$ to be the unit sensitivity counterpart to $\relu$.
\end{bond}

\begin{bond}[$\gelu$ and $\scaledgelu$]
The ``Gaussian error linear unit'' bond module $\gelu$ \cite{hendrycks2016gelu} is essentially a smoothed version of $\relu$. In any dimension $d$, $\gelu$ has inputs $\inputs\subset\R^d$, outputs $\outputs=\R^d$ and forward function
\begin{equation}
    \gelu\for : (x_1, \hdots, x_d) \mapsto (x_i \Phi(x_i))_{i=1,\hdots, x_d}
\end{equation}
where $\Phi(x) = \int_{-\infty}^x \tfrac{1}{\sqrt{2\pi}} e^{-t^2/2} dt$ is the cumulative distribution function of the standard Gaussian.

$\gelu$ is well-normed in the same sense as $\relu$. We similarly set $\scaledgelu = \sqrt{2} * \gelu$.
\end{bond}

\begin{bond}[$\meansub$]For any dimension $d$, the mean subtraction module $\meansub$ has inputs and outputs $\R^d$. It centers its input to have mean zero. The forward function is given by:
\begin{equation}
    \meansub\for : (x_1, \hdots, x_d) \mapsto (x_1 - \bar{\vx}, \hdots, x_d - \bar{\vx})
\end{equation}
and has sensitivity 1. It is well-normed, and since it is a linear mapping, it is 0-sharp.
\end{bond}

\begin{bond}[$\RMSdivide$]For any dimension $d$, the RMS division bond module $\RMSdivide$ has inputs and outputs $\R^d$. It normalizes its input to have unit RMS norm. The forward function is given by:
\begin{equation}
    \meansub\for : \vx \mapsto \frac{\vx}{\norm{\vx}_{\RMS}} = \frac{\sqrt{d} \, \vx}{\norm{\vx}_2}.
\end{equation}
and has sensitivity 1. While it is not automatically well-normed, as long as its inputs have $\norm{\vx}_{\RMS} \approx 1$, the required inequality is not very far off. Similarly, it is approximately $1$-sharp.
\end{bond}

\begin{bond}[$\layernorm$]For any positive integer $d$, the layer normalization bond module $\layernorm$ has inputs and outputs $\R^d$, and is just defined as the composition of modules
\begin{equation}
    \layernorm = \RMSdivide \circ \meansub.
\end{equation}
As with $\rmsdivide$, it is approximately well-normed and approximately $1$-sharp.
    
\end{bond}

\subsection{Module broadcasting}\label[subappendix]{app:broadcasting}

Let us briefly discuss a supplementary module operation, which we refer to as \emph{module broadcasting}.

\begin{definition}\label{def:broadcasting}
    Suppose $\module$ is a module with inputs $\inputs$, outputs $\outputs$ and weights $\weights$. Then for any $h \ge 1$, the $h$-times-broadcast of $\module$ is the module $\module^{(h)}$ with the same weight space $\weights$, mass, sensitivity and norm as $\module$, but inputs the Cartesian power $\inputs^h = \inputs \times \hdots \times \inputs$ and outputs $\outputs^h = \outputs \times \hdots \times \outputs$, and forward function
    \begin{equation}
        (\vw, (\vx_1, \hdots, \vx_h)) \mapsto (\module\for(\vw, \vx_1), \hdots, \module\for(\vw, \vx_h)).
    \end{equation}
    Since this is not defining a module with a new set of weights, we will usually just refer to the broadcast module by the same name $\module$, and consider this as just an extension of its forward function.
\end{definition}

For example, this allows us to define the action of linear modules $\linear(d_{\out}, d_{\inn})$ on inputs $\vx \in \R^{\ell \times d_{\inn}}$ to give outputs $\vy \in \R^{\ell \times d_{\out}}$, where $\ell$ is the context length parameter for a transformer (see \cref{app:attention}, \cref{app:transformer}, where it is also crucial for the construction of multi-headed attention). Additionally, one can view the basic $\absmodule$, $\relu$ and $\gelu$ modules as being broadcasts of the usual one-variable functions to take inputs and outputs in $\R^d$.

Let us briefly note:
\begin{proposition}\label{prop:broadcast_sens}
    If $\module$ is well-normed, then so is any broadcast of $\module$ taking $\inputs^h$ to $\outputs^h$, as long as the norms on $\inputs^h$ and $\outputs^h$ are taken to be either the ``mean $L^p$'' norms
    \begin{align}
        \norm{(\vx_1, \hdots, \vx_h)}_{\inputs^h} &=  \left(\frac{1}{h}(\norm{\vx_1}_{\inputs}^p + \hdots + \norm{\vx_h}_{\inputs}^p) \right)^{1/p} \\
        \norm{(\vy_1, \hdots, \vy_h)}_{\outputs^h} &=  \left(\frac{1}{h}(\norm{\vy_1}_{\outputs}^p + \hdots + \norm{\vy_h}_{\outputs}^p) \right)^{1/p}
    \end{align}
   for $1 \le p \le \infty$; when $p = \infty$ this is just the max norm.
   In the case that $\module$ is a bond module (so $\weights = 0$, any scalar multiple of the mean $L^p$ norm can be used (including the standard $L^p$ norm).
\end{proposition}

The situation for sharpness is a bit more complicated; we discuss this in \cref{app:broadcast_sharpness}.

\subsection{Mass taring}\label[subappendix]{app:taring}

In order to make working with the mass parameter of modules a bit easier, let us introduce an auxiliary operation:

\begin{definition}[Tare]\label{def:tare}
For any module $\module$ and positive real number $\mnew$, the module $\tare(\module, \mnew)$ has the exact same inputs, outputs and weights as $\module$; the same forward function, the same sensitivity and the same norm; but has mass
\begin{equation}
    \tare(\module, \mnew)\mass = \mnew.
\end{equation}
\end{definition}
This resets the mass of $\module$. If $\module$ is a compound module, one could also reset the masses of all its submodules, by taking $\tare(\module_k, \mnew * \tfrac{\module_k\mass}{ \module\mass})$ for every submodule $\module_k$, to ``reconstruct'' the computation graph for $\tare(\module, \mnew)$. 

This way, one can build complex modules starting from atomic modules with unit masses, and then using $\tare$ later to reset their masses to desired quantities for better feature learning with normed descent as in \cref{prop:character}.

\subsection{Compound modules and neural networks}\label[appendix]{app:network-design}

Composition, concatenation and the secondary operations of addition, scalar multiplication and iterated concatenation allow us to build a wide variety of neural networks which thus come automatically endowed with the modular norm.

Deep neural networks are typically built as long series of compositions. Let us introduce some terminology:

\begin{definition}[Blocks and deep networks]
A \emph{deep neural network} is a module $\module$ formed by a composition
\begin{equation}
    \module = \Final \circ \Block_L \circ \hdots \circ \Block_1 \circ \Initial
\end{equation}
where $\Initial, \Block_1, \hdots, \Block_L, \Final$ are modules; the number of blocks $L \ge 1$ is the \emph{depth} of the network.
\end{definition}

Typically, each of $\Block_1, \hdots, \Block_L$ will be copies of the same module (allowing them to take different weight values, of course), so that the network can be written as an iterated composition
\begin{equation}
    \module = \Final \circ \Block^L \circ \Initial.
\end{equation}
$\Initial, \Block, \Final$ can be principle be any module one likes, but usually $\Initial$ is often some form of embedding module, and $\Final$ is usually a linear module.

As for the form of $\Block$, we found the following design principle to be quite useful in practice:

\begin{center}
\emph{Arrange so that each $\Block$ has unit sensitivity.}
\end{center}

This ensures that the sensitivity of the whole network stays bounded as $L \to \infty$ (this will also be the case if we ensure that $\Block\lip = 1 + O(1/L)$, but unit sensitivity has the advantage that the modular norm becomes very explicit). With this in mind:

\begin{compound}[Residual network]
    Suppose that $\module$ is a module of unit sensitivity whose inputs and outputs are the same space $\inputs$. For any $L \ge 1$, consider the \emph{residual block}
    \begin{equation}
        \Block = \tfrac{L-1}{L} * \identity + \tfrac{1}{L} * \module
    \end{equation}
    and write $\residual{L}{\module} = \Block^L$. This is of unit sensitivity, well-normed if $\module$ is, and moreover by \cref{thm:sharpness_residual_modules} is sharp with O(1) sharpness if $\module$ is.
    
    A general \emph{residual network with residue $\module$} is any neural network of the form
    \begin{equation}
        \Final \circ \residual{L}{\module} \circ \Initial.
    \end{equation}
    In practice, we will want to apply one more operation: we will want to \emph{tare the mass of the residual blocks}. To this end, the \emph{residual network with residue $\module$, depth $L$ and total block mass $m > 0$ is}
    \begin{equation}
    \Final \circ \tare(\residual{L}{\module}, m) \circ \Initial.
    \end{equation}
\end{compound}

Let us give two basic example of residual networks.

\begin{compound}[ResMLP]
    This is a simple residual variation on the multi-layer perceptron. For a width $d \ge 1$, consider the unit sensitivity module
    \begin{equation}
        \module(d) = \meansub \circ \absmodule \circ \linear(d,d) \circ \RMSdivide.
    \end{equation}
    This particular order of operations is inspired by a reecent paper of \citet{yang2024tensor}.
    
    We invite the reader to compare this to something like $\relu \circ \linear(d,d) \circ \layernorm$: three core operations are being performed (but in a different order in both cases): the inputs are being normalized; the inputs are being centered; and the inputs are passed through a nonlinearity that mutates just the negative coordinates.

    The $\resmlp$ network has as its residue an iterated composition of $\module(d)$, where the number of copies of $\module(d)$ in each residue is called the \emph{block depth} and denoted $B$. It also has just linear initial and final modules. Thus the $\resmlp$ network of depth $L$, width $d$, block depth $B$ and total block mass $m > 0$ is
    \begin{equation}
        \resmlp = \linear(d_{\out}, d) \circ \tare(\residual{L}{\module(d)^B}, m) \circ \linear(d, d_{\inn})
    \end{equation}
    where $d_{\inn}$ is the number of features of the data, and $d_{\out}$ the desired number of output features of the network.
    
    Usually we suggest taking $B = 1$ or $2$, and $m \sim 1$.
\end{compound}

\begin{compound}[ResNet]
    This is a version of ResNet for image classification tasks. For a width $d \ge 1$ and kernel size $K$, consider similarly to above the unit sensitivity module
    \begin{equation}
        \module(d, K) = \meansub \circ \absmodule \circ \conv(d,d,K) \circ \RMSdivide.
    \end{equation}
    As in the $\resmlp$, the $\resnet$ network is a residual network with as its residue an iterated composition of $B$ copies of $\module(d,K)$ where $B$ is the \emph{block depth}. Its initial and final modules are given by
    \begin{align}
        \Initial &= \conv(d, c_{\inn}, K) \\
        \Final &= \linear(d_{\out}, d_{\mathrm{total}}) \circ \mathsf{AvgPool}
    \end{align}
    where $\mathsf{AvgPool}$ is an additional bond module implementing adaptative average pooling. Here, $c_{\inn}$ is the number of channel dimensions of the input image, $d_{\mathrm{total}} = d * H * W$ is the total dimension of the hidden representation, and $d_{\out}$ is the desired number of output features (note that in Modula we include an additional dummy module \python{Flatten} to change the tensor shape before passing through the final layer). The $\resnet$ network of depth $L$, width $d$, block depth $B$, kernel size $K$ and total block mass $m$ is thus:
    \begin{equation}
        \resnet = \Final \circ \tare(\residual{L}{\module(d,K)^B}, m) \circ \Initial.
    \end{equation}
    As defaults, we suggest taking $B = 2, K = 3$ and $m \sim 20$.

\end{compound}

\subsection{Case study I: Attention}\label[subappendix]{app:attention}

Let us now focus on the construction of a single \emph{multi-headed attention module} in this framework. The attention module should have, as both inputs and outputs, $\inputs = \R^{\ell \times d}$ where $\ell$ is the context length and $d$ is the embedding dimension. The attention module itself will depend on three additional dimensional arguments:
\begin{itemize}
    \item $h$, the number of heads;
    \item $d_Q$, the key/query dimension;
    \item $d_V$, the value dimension;
\end{itemize}
as well as an $\ell \times \ell$ matrix $\mask$, which we usually take to be either
\begin{equation}
    \mask_{ij} = \begin{cases}0 & \text{if } i \geq j \\
    -\infty & \text{otherwise}
    \end{cases}
\end{equation}
for causal attention, and $\mask = 0$ for non-casual attention.

The core of the attention module is a bond module which we call \emph{functional attention}.

\begin{bond}[$\funcattn$]\label{ex:attention}Take positive integers $\ell, d_Q, d_V$ and mask matrix $\mask$. The corresponding \emph{functional attention} is the bond module of unit sensitivity, inputs $\inputs = \R^{\ell \times d_Q} \times \R^{\ell \times d_Q} \times \R^{\ell \times d_V}$, outputs $\outputs = \R^{\ell \times d_V }$, and forward function
\begin{equation}\label{eq:funcattn}
    \funcattn\for(\vq, \vk, \vv) = \softmax\left(\frac{\vq \vk^\top}{d_Q} + \mask \right)\vv.
\end{equation}
Moreover, we set $\funcattn\lip = 1$.
\end{bond}

In theory, one could try break up attention further into constituent more basic modules (such as scaled dot product, softmax, etc), but keeping $\funcattn$ as the basic unit one to leverage efficient implementations of attention such as FlashAttention \cite{dao2022flashattention}.

In fact, a perhaps surprising result is that with the above $\frac{1}{d_Q}$ scaling of the dot product, we can estimate the sensitivity and sharpness of $\funcattn$. This relies on giving norms for the input and output spaces; these norms are chosen to be
\begin{equation}
    \norm{(\vq, \vk, \vv)}_{\inputs} = \norm{\vq}_{\inftyRMS} + \norm{\vk}_{\inftyRMS} + \norm{\vv}_{\inftyRMS}, \quad \norm{\vy}_{\outputs} = \norm{\vy}_{\inftyRMS}
\end{equation}
where $\norm{\cdot}_{\inftyRMS}$ is the \emph{infinity-RMS-norm} on $\R^{\ell \times d}$ defined from the standard root-mean-square norm $\norm{\cdot}_{\RMS}$ on $\R^d$ by
\begin{equation}
    \norm{\vx}_{\inftyRMS} := \max_{i = 1,\hdots, \ell} \norm{\vx_{i*}}_{\RMS}.
\end{equation}

\begin{proposition}\label{prop:attn-sens-sharpness}Over the space of inputs $\vq, \vk, \vv$ with each $\norm{\vq}_{\inftyRMS}, \norm{\vk}_{\inftyRMS}, \norm{\vv}_{\inftyRMS} \leq 1$, the functional attention module $\funcattn$ is well-normed, and moreover is sharp with sharpness constant $\gamma = 3$.
\end{proposition}

The proof is given in \cref{app:proofs}. We thus choose to adopt a $\frac{1}{d_Q}$-dot-product scaling in our implementation of attention-- a rigorous bound as above is not possible for $\frac{1}{\sqrt{d_Q}}$-scaling, for instance.

We can then immediately define a single head of attention.

\begin{compound}[Single-headed attention]For positive integers $\ell, d, d_Q, d_V$ and a choice of $\mask$, take four instances of the linear module, for the query, key, value and exit parameters:
\begin{align}
    \querymod &= \linear(d_Q, d) \\
    \keymod &= \linear(d_Q, d) \\ 
    \valuemod &= \linear(d_V, d) \\
    \exitmod &= \linear(d, d_V)
\end{align}
which by broadcasting we consider to have inputs of shape $\R^{\ell \times d}$. The single-headed attention $\shattn$ module is then the composition
\begin{equation}
    \shattn = \exitmod \circ \frac{1}{3} * \funcattn \circ (\querymod, \keymod, \valuemod).
\end{equation}
The scalar multiplication factor of $\frac{1}{3}$ ensures that $\shattn$ has unit sensitivity.
\end{compound}

For multiple heads of attention, we simply take advantage of module broadcasting (\cref{def:broadcasting}): 

\begin{compound}[Multi-headed attention]\label{com:mhattn}For positive integers $\ell, d, h, d_Q, d_V$ and a choice of $\mask$, take four instances of the linear module:
\begin{align}
    \querymod &= \linear(h * d_Q, d) \\
    \keymod &= \linear(h * d_Q, d) \\ 
    \valuemod &= \linear(h * d_V, d) \\
    \exitmod &= \linear(d, h * d_V)
\end{align}
which by broadcasting we consider to have inputs of shape $\R^{\ell \times d}$. The multi-headed attention $\mhattn$ module is then the composition:
\begin{equation}\label{eq:mhattn}
    \mhattn = \exitmod \circ \frac{1}{3} * \funcattn^{(h)} \circ (\querymod, \keymod, \valuemod)
\end{equation}
where $\funcattn$ is broadcast over the heads dimension. Note that in Modula, we do this by creating dummy bond modules called \python{AddHeads} and \python{RemoveHeads} to reshape the tensors and create/remove the explicit head dimension.

As in the single-headed case, the scalar multiplication factor of $\frac{1}{3}$ ensures unit sensitivity.
\end{compound}

\subsection{Case study II: GPT}\label[subappendix]{app:transformer}

Let us now build an auto-regressive transformer similar to GPT-2 \cite{GPT-2} or nanoGPT \cite{karpathy-nanogpt} in this framework. Fix positive integers $\ell, d, h, d_Q, d_V$ (usually $h$ divides $d$ and $d_Q = d_V = d/h$). In addition to \cref{com:mhattn} from the earlier, consider the 2-layer MLP:
\begin{equation}\label{eq:transformer_MLP}
    \mlp = \linear(d, 4d) \circ \sqrt{2} * \gelu \circ \linear(4d, d)
\end{equation}
where we are using the scalar correction so that $\gelu$ has unit sensitivity, and using module broadcasting so that it can take inputs and outputs $\R^{\ell \times d}$. Fix a depth $L \ge 1$, and consider the following two modules, whose input and output spaces are $\R^{\ell \times d}$:
\begin{align}
    \Block_{\mlp} &:= \tfrac{2L-1}{2L} * \identity + \tfrac{1}{2L} * \mlp \circ \layernorm_d \\
    \Block_{\attn} &:= \tfrac{2L-1}{2L} * \identity + \tfrac{1}{2L} * \mhattn \circ \layernorm_d
\end{align}
where $\layernorm_d$ refers to taking $\layernorm$ in the embedding dimension (i.e. the rows of matrices in $\R^{\ell \times d}$, as distinct from normalizing all $\ell \times d$ coordinates together). This can alternately be thought of as just taking the usual $\layernorm$ on $\R^d$ and broadcasting it to take inputs and outputs $\R^{\ell \times d}$.

Suppose that $N$ is the number of tokens. For the initial module, take two embeddings of the $N$ tokens and $\ell$ context positions
\begin{equation}
    \embed_{\mathrm{tok}} = \embed(N, d), \qquad \embed_{\mathrm{pos}} = \embed(\ell, d)
\end{equation}
and form the mass one, sensitivity one module
\begin{equation}\label{eq:transformer_initial}
    \Initial = \tare(\tfrac{1}{2} * \embed_{\mathrm{tok}} + \tfrac{1}{2} * \embed_{\mathrm{pos}}, 1).
\end{equation}
The final module is just
\begin{equation}
    \Final = \linear(N, d) \circ \layernorm_d.
\end{equation}
The depth $L \ge 1$, width $d$, total block mass $m > 0$ $\gpt$ module is thus
\begin{equation}
\gpt = \Final \circ \tare((\Block_{\mlp} \circ \Block_{\attn})^L, m) \circ \Initial.
\end{equation}
We suggest, as a default value, a total block mass of $m \sim 5$.
\clearpage
\section{More on Smoothness and Sharpness}\label[appendix]{app:sharpness}

\subsection{Underlying every estimate: The Gauss-Newton decomposition}
All our estimates of sharpness for compound modules, as well as the smoothness estimate \cref{thm:smoothness} for loss functions, depend on an application of the chain rule to compute second derivatives which in the optimization context is sometimes called the Gauss-Newton decomposition.

Namely, if $\vf : \R^{d_0} \to \R^{d_1}$ and $\vg : \R^{d_1} \to \R^{d_2}$, then the second derivative of their composite $\vh = \vg \circ \vf$ is computed by
\begin{equation}\label{eq:gauss-newton}
\vv \diamond \nabla^2 \vh \diamond \vw = (\nabla \vf \diamond \vv) \diamond \nabla^2 \vg \diamond (\nabla \vf \diamond \vw) + \nabla \vg \diamond (\vv \diamond \nabla^2 \vf \diamond \vw)
\end{equation}
for any $\vv, \vw \in \R^{d_0}$, or for short
\begin{equation}
    \nabla^2 \vh (\cdot, \cdot) = \nabla^2 \vg (\nabla \vf (\cdot), \nabla \vf(\cdot)) + \nabla \vg (\nabla^2 \vf(\cdot, \cdot)).
\end{equation}
Indeed, this amounts to simply the following expression for partial derivatives:
\begin{equation}
    \frac{\d^2 \vh}{\d x_i \d x_j}
    = \sum_{k, l} \frac{\d^2 \vg}{\d y_k \d y_l} \frac{\d f_k}{\d x_i} \frac{\d f_l}{\d x_j} + \sum_{k} \frac{\d \vg}{\d y_k} \frac{\d^2 f_k}{\d x_i \d x_j}.
\end{equation}

\subsection{Sharpness under composition and concatenation}
Here, we state the two formulae for computing the sharpness of a composition and a concatenation of two modules. The proofs are given in \cref{app:proofs}.

\begin{proposition}[Sharpness under composition]\label{prop:sharp_comp}
Suppose that $\module_2$ and $\module_1$ are well-normed, composable modules that are respectively $(\alpha_2, \beta_2, \gamma_2)$-sharp and $(\alpha_1, \beta_1, \gamma_1)$-sharp. Under the shorthand that $\smash{p_k \equiv \frac{\module_k\mass}{\module_1\mass + \module_2\mass}}$ and $\mu_k \equiv \module_k\lip$,
the composite $\module_2 \circ \module_1$ is $(\alpha, \beta, \gamma)$-sharp for:
\begin{align}
\label{eq:sharp_comp_alpha}    \alpha &= 
    \tfrac{1}{\mu_2} p_1^2 \alpha_1 + p_2^2 \alpha_2 + \tfrac{2}{\mu_2} p_1p_2 \beta_2 + \tfrac{1}{\mu_2^2} p_1^2 \gamma_2, \\
\label{eq:sharp_comp_beta}    \beta &= p_1 \beta_1 + \mu_1 p_2 \beta_2 + \tfrac{\mu_1}{\mu_2} p_1 \gamma_2, \\
\label{eq:sharp_comp_gamma}    \gamma &= \mu_2 \gamma_1 + \mu_1^2 \gamma_2.
\end{align}
\end{proposition}

\begin{proposition}[Sharpness under concatenation]\label{prop:sharp_concat}
Suppose that $\module_1$ and $\module_2$ are well-normed, concatenatable modules that are respectively $(\alpha_1, \beta_1, \gamma_1)$-sharp and $(\alpha_2, \beta_2, \gamma_2)$-sharp. Under the shorthand that $\smash{p_k \equiv \frac{\module_k\mass}{\module_1\mass + \module_2\mass}}$ and $\mu_k \equiv \module_k\lip$,
the tuple $(\module_1, \module_2)$ is $(\alpha, \beta, \gamma)$-sharp for:
\begin{align}
    \alpha &= 
    p_1^2 \alpha_1 + p_2^2 \alpha_2, \\
    \beta &= p_1 \beta_1 + p_2 \beta_2, \\
    \gamma &= \gamma_1 + \gamma_2.
\end{align}
\end{proposition}
Taken together, \cref{prop:sharp_comp,prop:sharp_concat} specify a recursive procedure for computing the sharpness of any compound module that is built from a set of well-normed modules of known sharpness.

\begin{remark}
    These two sets of formulas are actually \emph{associative}, as the reader may verify using their favorite computer algebra package. This means, for instance, that if $\module_1, \module_2, \module_3$ are successively composable, well-normed and each $(\alpha_k, \beta_k, \gamma_k)$-sharp, then the two sets of sharpness estimates coming from applying the above formulas for $\module_3 \circ (\module_2 \circ \module_1)$ and $(\module_3 \circ \module_2)\circ\module_1$ actually coincide. 
\end{remark}

\subsection{Sharpness under module broadcasting}\label[appendix]{app:broadcast_sharpness}

Suppose $\module$ is a well-normed module with inputs $\inputs$, outputs $\outputs$ and weights $\weights$, and suppose moreover that it is $(\alpha, \beta, \gamma)$-sharp. The broadcast module $\module^{(h)}$ has the same weights, mass, sensitivity and norm, but takes $\inputs^h$ to $\outputs^h$.

By \cref{prop:broadcast_sens}, $\module^{(h)}$ is well-normed, as long as the norms on $\inputs^h$ and $\outputs^h$ are taken to be
    \begin{align}
        \norm{(\vx_1, \hdots, \vx_h)}_{\inputs^h} &= S* \left(\norm{\vx_1}_{\inputs}^p + \hdots + \norm{\vx_h}_{\inputs}^p \right)^{1/p} \\
        \norm{(\vy_1, \hdots, \vy_h)}_{\outputs^h} &= S* \left(\norm{\vy_1}_{\outputs}^p + \hdots + \norm{\vy_h}_{\outputs}^p \right)^{1/p}
    \end{align}
for $1 \leq p \leq \infty$; unless $\module$ is a bond module (and thus weight-less), we must take $S = h^{-1/p}$, otherwise $S$ can be any positive scalar.

A natural question is whether $\module^{(h)}$ is also sharp, and if so what its sharpness constants are, with respect to these norms. More or less the same proof as for \cref{prop:broadcast_sens} shows that the $\alpha$ and $\beta$ bounds for sharpness are always true, with the same $\alpha, \beta$. The $\gamma$ bound is trickier however, and depends subtly on the chosen $S, p$. We highlight three cases where one can say something interesting.

\emph{Case 1. $p = \infty, S = 1$}. For the $L^{\infty}$ norm, we have that $\module^{(h)}$ is $(\alpha, \beta, \gamma)$-sharp with the same $\alpha, \beta, \gamma$ by a more or less immediate proof.

\emph{Case 2. $p < \infty, S = 1$}. For the ``standard'' $L^p$-norms, we have that $\module^{(h)}$ is $(\alpha, \beta, \gamma)$-sharp with the same $\alpha, \beta, \gamma$. The proof is direct, using the inequality
\begin{equation}
    (x_1^p \widetilde{x}_1^p + \hdots + x_h^p \widetilde{x}_h^p)^{1/p} \leq (x_1^p + \hdots + x_h^p)^{1/p}(\widetilde{x}_1^p + \hdots + \widetilde{x}_h^p)^{1/p}
\end{equation}
for any positive reals $x_1, \hdots, x_h, \tilde{x}_1, \hdots, \tilde{x}_h$; however this is \emph{a very weak inequality} and so leads to very pessimistic sharpness estimates for large $h$.

\emph{Case 3. $p = 2, S = 1/\sqrt{h}$}. This is the ``RMS norm'' case. As in Case 2, one could use a very weak inequality to obtain the pessimistic result that $\module^{(h)}$ is $(\alpha, \beta, \sqrt{h} * \gamma)$-sharp. However, one could also make the following observation: if $h$ is large, and $x_1, \hdots, x_h$ are sampled from any normal distribution $N(\mu, \sigma^2)$, then
\begin{equation}\label{eq:L4_L2_comparison}
    \left(\frac{1}{h}(x_1^4 + \hdots + x_h^4)\right)^{1/2}
    \approx 
    \sqrt{3} \left(\frac{1}{h}(x_1^2 + \hdots + x_h^2)\right).
\end{equation}
In particular, this justifies the statement that ``for large $h$, the broadcast module $\module^{(h)}$ is approximately $(\alpha, \beta, \sqrt{3} * \gamma)$-sharp''. While in actual deep learning contexts, the assumption that $x_1, \hdots, x_h$ are sampled from a normal distribution may not be valid, one should still expect the ratio between the two sides of \cref{eq:L4_L2_comparison} to stay O(1) as $h \to \infty$, and so even if the ``$\sqrt{3}$ rule'' is insufficient, the effective sharpness of the broadcast module should not blow up as $h \to \infty$.

\subsection{Smoothness estimates for common error measures} \label[subappendix]{app:errors}

Suppose $\ell : \outputs \times \targets \to \R$ is an error measure. In \cref{thm:smoothness}, we showed that smoothness estimates on $\ell$ together with sharpness of a neural network imply smoothness of the corresponding average error loss function. The precise estimates are that $\ell$ is $\sigma$-Lipschitz and $\tau$-smooth in the module output, in the sense that:
\begin{align}\abs{\nabla_{\vy} \ell(\vy, \vt) \diamond \D \vy} &\leq \sigma \, \norm{\D \vy}_{\outputs} &&\textnormal{for all } \D \vy \in \outputs \textnormal{ and } \vt \in \targets;\\
\abs{\D \vy \diamond \nabla^2_{\vy \vy} \ell(\vy,\vt) \diamond \D \widetilde{\vy}} &\leq \tau \, \norm{\D \vy}_{\outputs}\,\norm{\D \widetilde{\vy}}_{\outputs} &&\textnormal{for all } \D \vy, \D \widetilde{\vy} \in \outputs \textnormal{ and } \vt \in \targets.
\end{align}
We now present estimates on $\sigma$ and $\tau$ for square and cross-entropy error. Both estimates will be in terms of \emph{the value of the average loss function $\el$ itself}, rather than being truly global over the entire output space $\outputs$. Thus, to apply them to real learning problems, one should \emph{measure} the average loss $\el$ at initialization, and use this for estimates for $\sigma$ and $\tau$; we are implicitly making the assumption that under gradient descent the loss decreases.

\subsection*{Square error}

Consider square error for a $d$-class classification problem. Thus, $\outputs = \R^d$ and $\targets = \{1,\hdots,d\}$. Consider the RMS norm on $\outputs$, and define the error function
\begin{equation}
    \ell(\vy, t) = \frac{1}{2d}\left(y_1^2 + \hdots + (y_t - \sqrt{d})^2 + \hdots + y_d^2\right) \qquad \text{for } \vy, t \in \outputs \times \targets.
\end{equation}
(the slightly non-standard scalings are due to the choice of RMS norm on $\outputs$). The first and second partial derivatives of $\ell$ are given by
\begin{equation}
    \frac{\d \ell}{\d y_i}(\vy,t) = \frac{1}{d} (y_i - \delta_{it} \sqrt{d}), \qquad
    \frac{\d^2 \ell}{\d y_i \d y_j}(\vy, t) = \frac{1}{d} \delta_{ij}
\end{equation}
The desired constants $\sigma, \tau$ can then be computed as maxima:
\begin{equation}
    \sigma = \max_{\norm{\vz}_{\RMS}=1} \sum_i \frac{\d \ell}{\d y_i} z_i, \qquad
    \tau = \max_{\norm{\vz}_{\RMS}=1} \sum_{i,j} \frac{\d^2 \ell}{\d y_i \d y_j} z_i z_j
\end{equation}
which from the above formulas amounts exactly to
\begin{equation}
    \sigma = \sqrt{\ell(\vy, t)}, \qquad \tau = 1.
\end{equation}
To translate this into a bound for the average loss function $\el$, note that square root is a \emph{concave} function. Thus if we have outputs $\vy_1, \hdots, \vy_B$ with true classes $t_1, \hdots, t_B$, Jensen's inequality yields
\begin{equation}
    \frac{1}{B}\sum \sqrt{\ell(\vy_b, t_b)}
    \leq \sqrt{\tfrac{1}{B} \sum \ell(\vy_b, t_b)} = \sqrt{\el}
\end{equation}
allowing us to use $\sigma = \sqrt{\el}$ as our estimate for \cref{thm:smoothness}.

\subsection*{Cross-entropy error}
Consider cross-entropy error for a $d$-class classification problem. Thus, $\outputs = \R^d$ and $\targets = {1, \hdots, d}$. For $\vy \in \R^d$ and $t \in \targets$, write
\begin{equation}p_t(\vy) = \frac{e^{y_t}}{\Sigma_j e^{y_j}}
\end{equation}
and consider the error function
\begin{equation}
    \ell(\vy, t) = - \log(p_t(\vy)).
\end{equation}
The first and second partial derivatives of $\ell$ are given by
\begin{equation}
        \frac{\d \ell}{\d y_i}(\vy, t) = p_i - \delta_{it}, \qquad 
    \frac{\d^2 \ell}{\d y_i y_j}(\vy, t) = \delta_{ij} p_i - p_i p_j.
\end{equation}
Consider again the RMS norm on $\outputs$. An estimate on $\sigma$ can thus be computed as
\begin{equation}
\max_{\norm{\vz}_{\RMS}=1} \sum_i \frac{\d \ell}{\d y_i} z_i
= \sqrt{d} \left(\sum_i (p_i - \delta_{it})\right)^{1/2} \leq
\sqrt{d} * \sqrt{\ell}
\end{equation}
using the basic fact that if $p_1, \hdots, p_d$ are non-negative numbers that sum to 1, then
\begin{equation}
    p_1^2 + \hdots + (p_t - 1)^2 + \hdots + p_d^2 \leq -\log(p_t).
\end{equation}
(Indeed, for fixed $p_t$, the left hand side is maximized at $p_1 = 1-p_t$ and all other $p_i$ = 0; one then easily checked that $2(p-1)^2 \leq -\log(p)$ for all $0 < p \leq 1$.)

A similar concavity argument to the square error case then enables us to use $\sigma = \sqrt{d} * \sqrt{\el}$ as the first derivative bound for average cross-entropy loss.

The second derivative bound depends on more subtle information geometry. Indeed, $\tau$ can be computed to be
\begin{equation}
    \tau = d * \lambda
\end{equation}
where $\lambda$ is the largest eigenvalue of the matrix $\diag(\vp) - \vp \vp^T$. It is possible for this eigenvalue to be quite large (for instance, if $p_1 = p_2 = 1/2$ and all other $p_i = 0$, then $\lambda = 1/2$). However, the average eigenvalue is
\begin{equation}
    \frac{1}{d}\left(1 - \sum p_i^2\right) \le \frac{d-1}{d^2} < \frac{1}{d}.
\end{equation}
If we presumed that, in the course of a gradient descent optimizing the weights of a module $\module$, the output perturbations $\nabla \module \diamond \D \vw$ are only generically aligned with the eigenvectors of $\diag(\vp) - \vp \vp^T$, then we could use the ``effective'' smoothness bound $\tau = 1$.

Perhaps this is a dubious assumption however. A more conservative, but perhaps still dubious, assumption comes from assuming that the logits $\vy$ have roughly $N(0,1)$ entries---at least this could be more or less true at initialization. In this case, the largest eigenvalue $\lambda$ is with high probability bounded as
\begin{equation}
    \lambda \leq 1/\sqrt{d}
\end{equation}
justifying ``approximate'' smoothness bound of $\tau = \sqrt{d}$.
\clearpage
\section{Experimental Details}\label[appendix]{app:experiments}

\subsection{Datasets}
\label[subappendix]{sec:datasets}

All experiments with ResMLP and ResNet~\cite{He2015DeepRL} are done with the CIFAR-10~\cite{Krizhevsky09learningmultiple} image dataset with standard train and test splits. For data augmentation on the training set, we use random crop, random horizontal flip and PyTorch AutoAugment.

For the GPT~\cite{GPT-2} transformer experiments, we compared three different datasets:
\begin{enumerate}[label=(\alph*)]
\item The Shakespeare corpus, using character-level tokens~\cite{karpathy2022tiny};
\item The TinyStories database \cite{tinystories} using sub-word level tokenization;
\item OpenWebText using sub-word level tokenization~\cite{Gokaslan2019OpenWeb}.
\end{enumerate}
No data augmentation was used on the language data. We used data splitting code from \citep{karpathy-nanogpt}.

\subsection{Architectures}
\label[subappendix]{sec:architectures}

Full details of the ResMLP, ResNet and GPT architectures we used are detailed in Appendices \cref{app:network-design} and \cref{app:transformer}. In every experiment, we used:
\begin{enumerate}[label=(\alph*)]
\item cross-entropy loss with no weight decay;
\item block depth $B = 2$ for ResMLP and ResNet;
\item kernel size $K = 3$ for ResNet;
\item $h = 8$ heads for GPT, with query and value dimensions $d_Q = d_V = d/h$ where $d$ is the embedding dimension (width);
\item context length $128$ for GPT, except for the experiment in \cref{app:context_length}.
\end{enumerate}

\begin{figure}
    \centering
    \includegraphics[width=\linewidth]{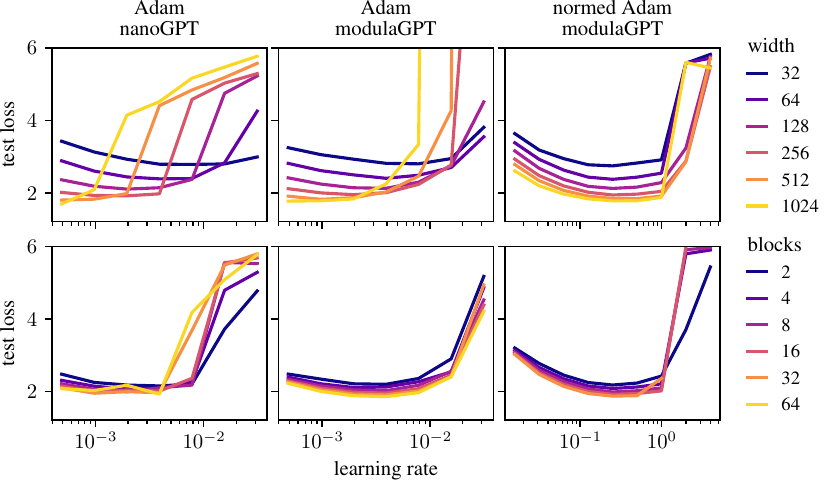}
    \mycaption{Comparing to a standard transformer implementation.}{Since we used our own well-normed GPT implementation for the experiments in this paper (here referred to as modulaGPT) we wanted to check its performance was on par with a standard nanoGPT implementation. These plots show learning rate sweeps for varying width and depth for Adam on nanoGPT, as well as Adam and normed Adam on modulaGPT. Even without normed updates, the architectural changes and orthogonal initialization used in Modula seem to already improve transfer compared to nanoGPT.}
    \label{fig:nanogpt-vs-modular}
\end{figure}

\subsection{Hardware}\label[subappendix]{sec:hardware}
All experiments were run on NVIDIA GPUs using \texttt{float32}-precision. We used a combination of \texttt{TITAN-RTX}, \texttt{RTX-3090}, \texttt{V100}, \texttt{Ada6000}, and \texttt{H100} devices. Each data point in the experiments takes up to $5$ hours, depending on the computing device used. We ran over 1000 training runs in total.

\subsection{Comparing to standard nanoGPT architecture}\label[subappendix]{app:std_comparison}

Our implementation of GPT in \python{Modula} has certain differences from off-the-shelf architectures such as nanoGPT \cite{karpathy-nanogpt}. We would summarize the overall changes to transformer architecture and training the following three points:
\begin{enumerate}[label=(\Roman*)]
\item the mathematical \emph{architecture has slightly different coefficients};
\item we initialize weight matrices to be \emph{orthogonal rather than Gaussian};
\item we train using \emph{normalized weight updates}.
\end{enumerate}

\begin{figure}
    \centering
    \includegraphics[width=\linewidth]{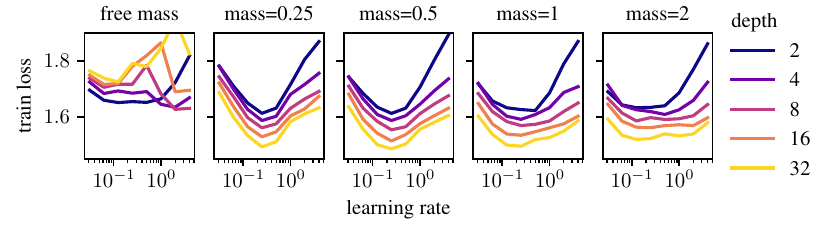}
    \mycaption{Comparing mass allocation strategies.}{We train a ResMLP with width 64 and 2 layers per block on CIFAR-10. In the first sub-plot titled ``free mass'', we set every atomic module to have unit mass, so that as depth is scaled the masses of the input and output layer become insignificant relative to the total mass of the hidden layers. In the other four subplots, we tare the total mass of the hidden layers to the value indicated in the subplot title. As can be seen, the taring strategy seems to work much better than the free mass strategy. So, at least in this experiment, it is good to keep a constant fraction of learning in the input and output layers even as depth is scaled.}
    \label{fig:mass-sweep-resmlp-appendix}
\end{figure}

The architectural choices we made were entirely informed by the desire for the network to be well-normed and have unit sensitivity: in particular this means that the network enjoys favorable signal propagation properties. In the language of modules, these architectural changes can be summarized as:
\begin{enumerate}[label=(\alph*)]
\item Each residual block in our architecture is of the form
\begin{equation}\tfrac{2L-1}{2L} * \identity + \tfrac{1}{2L} * \Block\end{equation}
where $\Block = \Block_{\mlp}$ or $\Block_{\attn}$, compared to $\identity + \frac{1}{\sqrt{L}} \Block$ suggested for nanoGPT;

\item We use a scaled dot product attention with $\frac{1}{d_Q}$ scaling, rather than $\frac{1}{\sqrt{d_Q}}$;

\item The forward function of our $\linear$ and $\embed$ modules (see \cref{sec:atoms}) includes scale factors $\sqrt{d_{\out}/d_{\inn}}$ and $\sqrt{d}$ respectively. 

\item We use several additional scalar multiplications to keep the network of unit sensitivity:
\begin{itemize}
    \item Each Attention module (\ref{eq:mhattn}) has a scalar factor of $\frac{1}{3}$;
    \item Each MLP module (\ref{eq:transformer_MLP}) has a scalar factor of $\sqrt{2}$;
    \item The token and position embeddings (\ref{eq:transformer_initial}) have a scalar factor of $\frac{1}{2}$.
\end{itemize}
\end{enumerate}

In \cref{fig:nanogpt-vs-modular}, we ran a comparison of the performance of the standard (unnormed) Adam optimizer trained on OpenWebText with:
\begin{enumerate}
    \item the nanoGPT architecture with Gaussian initialization;
    \item our implementation of GPT with orthogonal initialization.
\end{enumerate}
We found that even without using the normed optimizer, our implementation with orthogonal initialization \emph{transferred learning rate better}. We suggest that even the base Adam optimizer benefits from the above architectural changes.

\subsection{Full sweeps}\label[subappendix]{app:full-sweep}

\begin{figure}
    \centering
    \includegraphics[width=\textwidth]{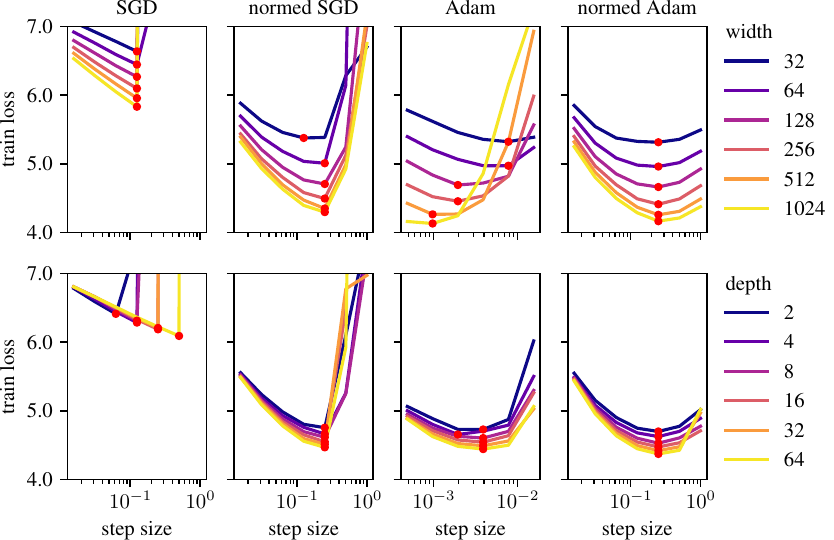}\vspace{5ex}
    \includegraphics[width=\textwidth]{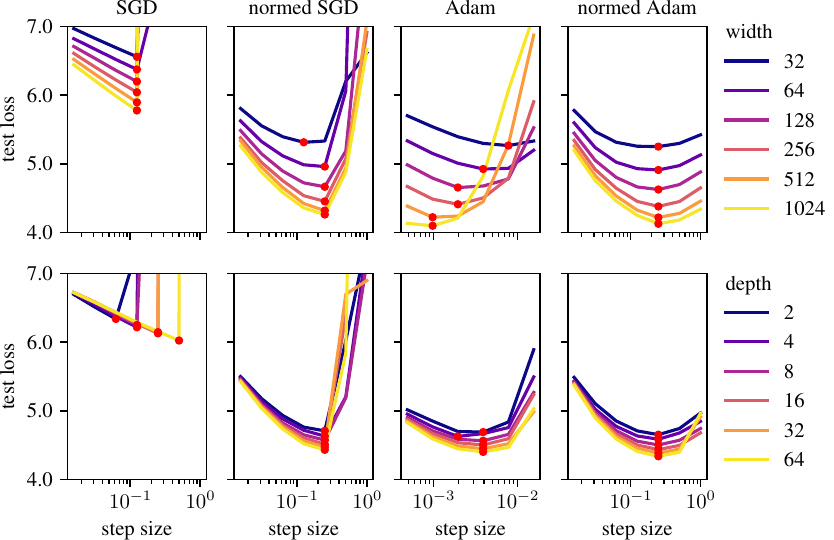}
    \mycaption{Learning rate transfer for GPT on OpenWebText.}{Training is done for 10k steps, at batch size 128, with SGD, Adam, and their normed versions. The total block mass for normed SGD/Adam is $m = 5$. Width scaling experiments are done at fixed depth 3, and depth scaling experiments are done at fixed width 128.}
    \label{fig:sweep-gpt}
\end{figure}

\begin{figure}
    \centering
    \includegraphics[width=\textwidth]{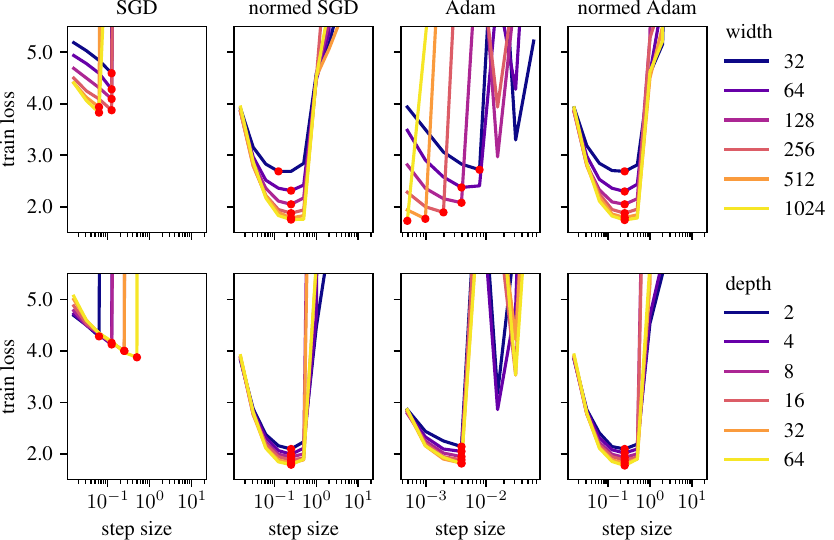}\vspace{5ex}
    \includegraphics[width=\textwidth]{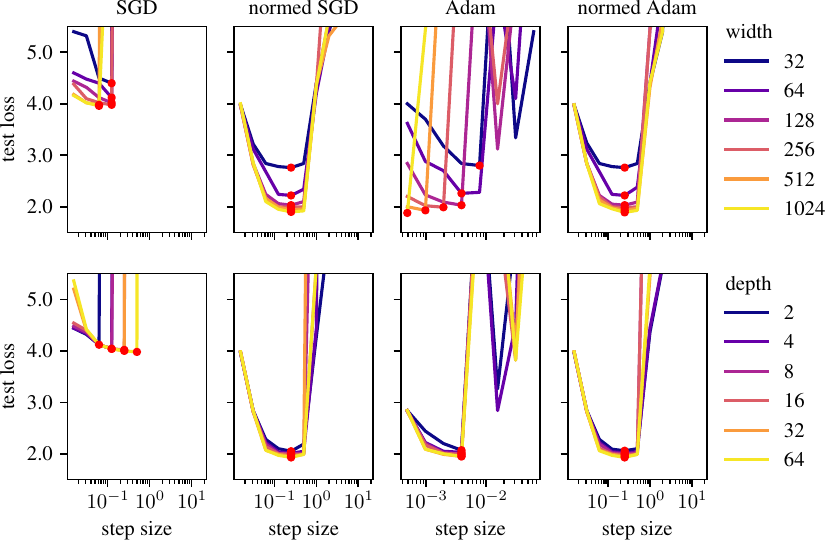}
    \mycaption{Learning rate transfer for GPT on TinyStories.}{Training is done for 10k steps, at batch size 128, with SGD, Adam, and their normed versions. The total block mass for normed SGD/Adam is $m = 5$. Width scaling experiments are done at fixed depth 3, and depth scaling experiments are done at fixed width 128.}
    \label{fig:sweep-gpt-tiny}
\end{figure}

\begin{figure}
    \centering
    \includegraphics[width=\textwidth]{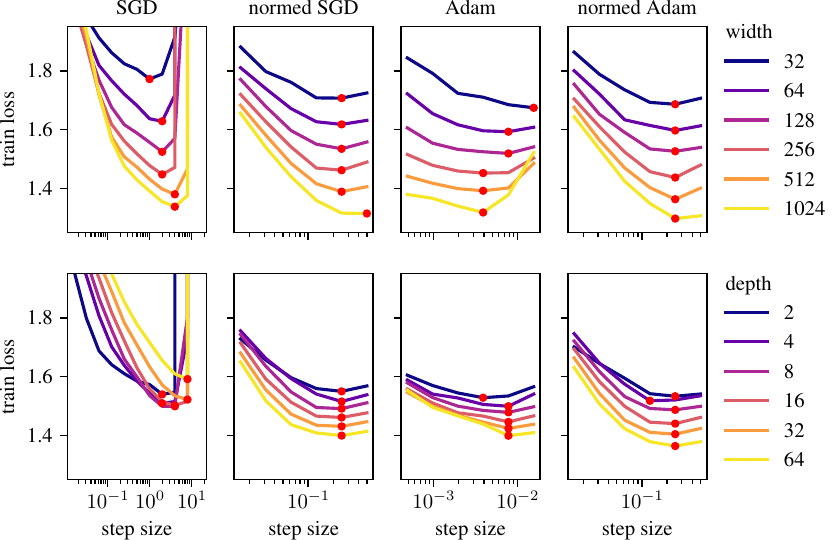}\vspace{5ex}
    \includegraphics[width=\textwidth]{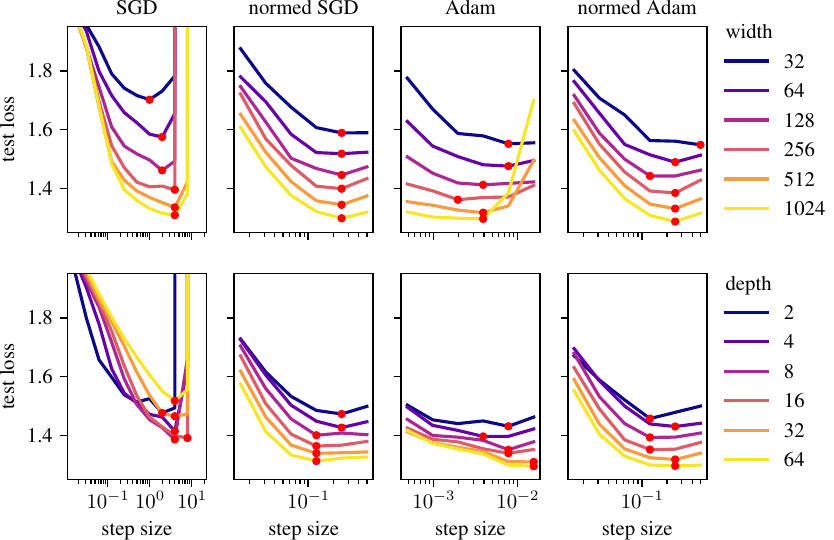}
    \mycaption{Learning rate transfer for ResMLP.}{ResMLP architectures on CIFAR-10 are trained for 10k steps, at batch size 128, with SGD, Adam, and their normed versions. The total block mass for normed SGD/Adam is $m = 1$. Width scaling experiments are done at fixed depth 3, and depth scaling experiments are done at fixed width 128.}
    \label{fig:sweep-mlp}
\end{figure}

\begin{figure}
    \centering
    \includegraphics[width=\textwidth]{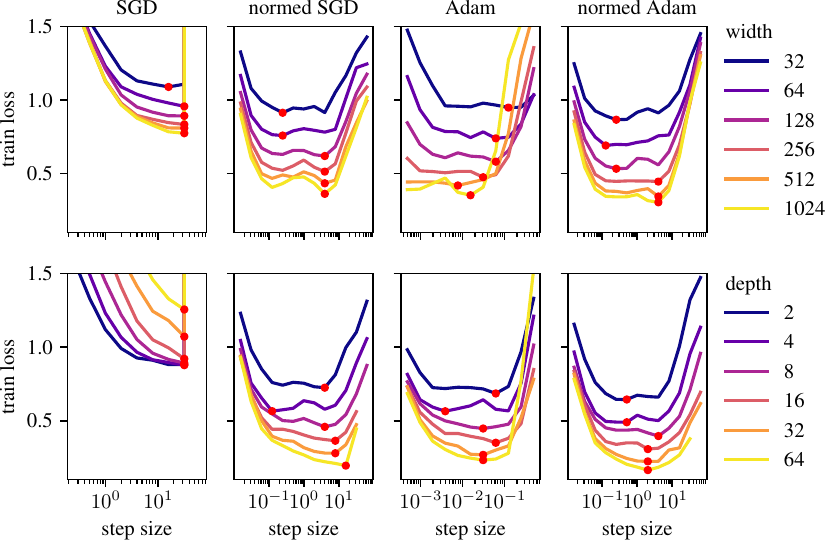}\vspace{5ex}
    \includegraphics[width=\textwidth]{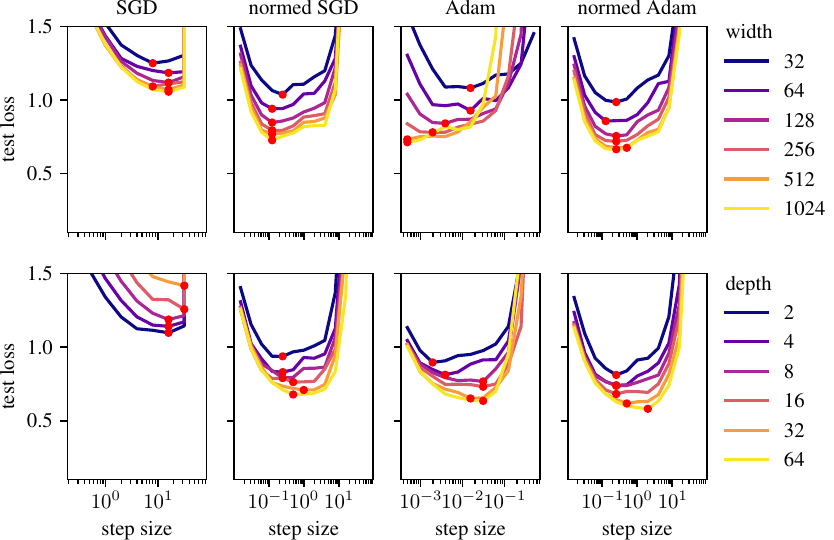}
    \mycaption{Learning rate transfer for ResNet.}{ResNet architectures on CIFAR-10 are trained for 10k steps, at batch size 128, with SGD, Adam, and their normed versions. The total block mass for normed SGD/Adam is $m = 20$. Width scaling experiments are done at fixed depth 3, and depth scaling experiments are done at fixed width 128.}
    \label{fig:sweep-cnn}
\end{figure}

In \cref{fig:sweep-gpt,fig:sweep-gpt-tiny,fig:sweep-mlp,fig:sweep-cnn}, at the end of this Appendix, we report on full learning rate sweep experiments, across width and depth, for GPT on OpenWebText and TinyStories, and ResMLP, ResNet on CIFAR-10.

We consistently find that the normed Adam optimizer matches or outperforms unnormed Adam in both test and training loss, all the while exhibiting significantly better transfer across width. The difference in depth transfer is less stark, however we posit that, in part, unnormed Adam is already benefiting from architectural changes we made to improve depth scaling.

Notice too that normed SGD consistently significantly outperforms ordinary SGD, often coming close to or matching the performance of Adam. We would like to highlight this, since SGD has a significantly lower memory requirement than Adam, and does not require any tuning of $\beta_2$.

\begin{figure}
    \centering
    \includegraphics[width=\linewidth]{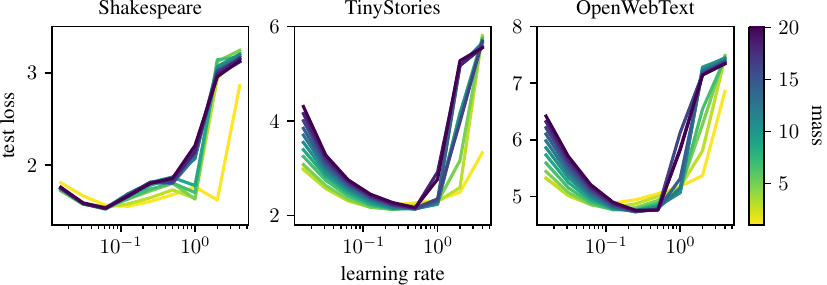}
    \mycaption{Mass and learning rate sweeps across datasets of increasing difficulty.}{A small GPT architecture of width 128 and 3 transformer blocks was trained on the Shakespeare, TinyStores and OpenWebText datasets. We varied the learning rate as well as the total mass of the blocks. Optimal mass and learning rate seem to transfer reasonably well from TinyStories to OpenWebText, and less well from the much smaller Shakespeare dataset.}
    \label{fig:mass-sweep-gpt-tasks}
\end{figure}

\subsection{Mass allocation}\label[subappendix]{app:mass_allocation}

A novel feature of our normed optimization framework is the need to choose a \emph{mass} parameter for each atomic module. In the context of networks of the form
\begin{equation}
    \mathsf{Network} = \mathsf{OutputLayer} \circ \mathsf{HiddenLayers} \circ \mathsf{InputLayer}
\end{equation}
where $\mathsf{HiddenLayers} = \Block^L$. We typically do this by assuming that $\Initial, \Final$ have mass 1, and by hand resetting the mass of $\mathsf{HiddenLayers}$ to be a fixed total mass $m > 0$, by calling $\tare(\mathsf{HiddenLayers}, m)$.

In this Appendix, we explore some different aspects the choice of $m$.

First, we tested whether or not calling $\tare$ is necessary in the first place. Not using tare would leave the ``free mass'' of $\mathsf{HiddenLayers}\mass = L * \Block\mass$; accordingly as $L$ grows large, the feature learning allotment (see \cref{prop:character}) for $\Initial$ and $\Final$ would grow smaller. Indeed, as the reader can see in \cref{fig:mass-sweep-resmlp-appendix}, this ``free mass'' arrangement for a ResMLP network on CIFAR-10, allowing the mass of $\mathsf{HiddenLayers}$ to grow with $L$ is very undesirable, and for good learning rate transfer with depth we must fix a mass.

The mass $m$ is thus left as a tunable parameter. We then tested the transferability of mass tuning. Specifically, we wanted to know:
\begin{enumerate}
    \item whether one can tune $m$ on a network of small width/depth, and expect that same $m$ to be close to optimal on a larger network;
    \item whether learning rate transfer across width/depth is itself dependent on selecting a good mass $m$;
    \item how sensitive the tuning for $m$ is: if there is a broad range of acceptable masses, or certain precise values lead to big improvements in train or test loss.
\end{enumerate}

Figures \cref{fig:mass} and \cref{fig:mass-sweep-resmlp-appendix} answer Question 1 above in the affirmative, in the context of ResMLP on CIFAR-10 and GPT on OpenWebText. Moreover, in the context of ResMLP on CIFAR-10, they give an answer of Question 2 and Question 3: learning rate transfer occurs at a range of values of $m$.

\cref{fig:mass-sweep-gpt-tasks} address Question 3 in the context of transformers, on three different datasets. Across all three datasets, a mass in the region $m \sim 5$ to $10$ is reasonable.

\subsection{Context length}\label[subappendix]{app:context_length}

\begin{figure}
    \centering
    \includegraphics[width=0.5\linewidth]{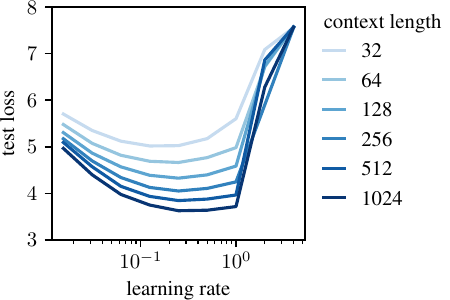}
    \mycaption{Context length transfer.}{We trained GPTs of various context lengths using normed Adam. As can be seen, learning rate transferred quite well across context length.}
    \label{fig:gpt-owt-context}
\end{figure}

Additionally, we also tested the dependence of the optimal learning rate for GPT training on OpenWebText on \emph{the context length}; the results are in \cref{fig:gpt-owt-context} Interestingly, we report good transfer of the optimal learning rate from small contexts to long contexts.

\subsection{Full sweep results}

The next four pages of the Appendix list results of our full learning rate sweeps over width/depth for GPT on OpenWebText and TinyStories, and ResMLP, ResNet on CIFAR-10.

\section{Proofs}\label[appendix]{app:proofs}

\subsection*{\cref{prop:character}: Feature learning is apportioned by mass}
\label{sec:massproof}

To prove \cref{prop:character}, it suffices to induct on the construction of a compound module $\module$ by composition and concatenation, with the atomic modules (where the inequality is just part of well-normed-ness) as the base case.

Indeed, suppose either $\module = \module_2 \circ \module_1$ or $\module = (\module_1, \module_2)$. Suppose $\vw_k$ is a weight for one of the atomic modules of $\module$, and write $m$ for the mass of this atomic module. Then $\vw_k$ is must be a weight of either $\module_1$ or $\module_2$; the inductive assumption is that
\begin{equation}
    \norm{\nabla_{\vw_k} \module_i \diamond \D \vw_k} \leq
    \frac{m}{\module_i\mass} * \norm{\D \vw}_{\module_i}
\end{equation}
where $i = 1$ or $2$ accordingly.

\emph{Case 1}. $\module = \module_2 \circ \module_1$ and $\vw_k$ is a weight of $\module_1$. From the chain rule we then must have:
\begin{align}
    \norm{\nabla_{\vw_k} \module \diamond \D \vw_k}
    &= \norm{\nabla_{\vx} \module_2 \diamond \nabla_{\vw_k} \module_1 \diamond \D \vw_k} \\
    &\leq \module_2\lip * \norm{\nabla_{\vw_k} \module_1 \diamond \D \vw_k} \quad \text{by well-normed-ness} \\
    &\leq \module_2\lip * \frac{m}{\module_1\mass} * \norm{\D \vw}_{\module_1} \quad \text{by assumption}\\
    &\leq \frac{m}{\module\mass} \norm{\D \vw}_{\module}
\end{align}
where the last line is by the definition of the norm of module composition.

\emph{Case 2}. $\module_2 \circ \module_1$ and $\vw_k$ is a weight of $\module_2$. The chain rule is not needed in this case, and we proceed straight from the inductive assumption:
\begin{align}
    \norm{\nabla_{\vw_k} \module \diamond \D \vw_k}
    &= \norm{\nabla_{\vw_k} \module_2 \diamond \D \vw_k}\\
    &\leq \frac{m}{\module_2\mass} * \norm{\D \vw}_{\module_2} \\
    &\leq \frac{m}{\module\mass} \norm{\D \vw}_{\module}.
\end{align}

\emph{Case 3}. $\module = (\module_1, \module_2)$. Given the symmetric roles of $\module_1, \module_2$, without loss of generality assume $\vw_k$ is a weight of $\module_1$. Then,
\begin{align}
    \norm{\nabla_{\vw_k}\module \diamond \D \vw_k}
    &= \norm{\nabla_{\vw_k} \module_1 \diamond \D \vw_k}\\
    &\leq \frac{m}{\module_1\mass} * \norm{\D \vw}_{\module_1} \\
    &\leq \frac{m}{\module\mass} \norm{\D \vw}_{\module}.
\end{align}
This completes the proof.

\subsection*{\cref{thm:sharpness_residual_modules}: Sharpness of residual networks}
\label{sec:residual-sharpness}
\label{app:res-proof}
Suppose $\module$ is a well-normed module of unit sensitivity on $(\inputs, \inputs, \weights)$ and is $(\alpha, \beta, \gamma)$-sharp. Then, by Proposition \ref{prop:sharp_comp} for any $L \ge 1$, the module $\tfrac{1}{L} * \module$ is well-normed, sensitivity $\tfrac{1}{L}$, and $(L \alpha, \beta, \tfrac{1}{L}\gamma)$-sharp.

The module $\tfrac{L-1}{L} * \identity$ is also well-normed, sensitivity $\tfrac{L-1}{L}$, and $(0,0,0)$-sharp. In particular, the sum
\begin{equation}
    \module_{res} = \tfrac{L-1}{L} * \identity + \tfrac{1}{L} * \module
\end{equation}
is well-normed, unit sensitivity, and $(L \alpha, \beta, \tfrac{1}{L}\gamma)$-sharp; it has the same mass as the original module.

We induct on the statement for $k = 1, 2, \hdots$ that $\module_{res}^k$ is $(\alpha_k, \beta_k, \gamma_k)$-sharp where
\begin{align}
    \alpha_k &= \frac{L}{k} \alpha + \frac{2\left(1 + 2 + \hdots + (k-1)\right)}{k^2}  \beta + \frac{\left(1^2 + 2^2 + \hdots + (k-1)^2\right)}{L k^2} \gamma \\
    \beta_k &= \beta + \frac{\left(1 + 2 + \hdots + (k-1) \right)}{Lk}   \gamma \\
    \gamma_k &=  \frac{k}{L} \gamma.
\end{align}
The base case is clearly true, and given the statement for $\module_{res}^k$, which has exactly $k$ times the mass as $\module_{res}$, we see that $\module_{res}^{k+1} = \module_{res} \circ \module_{res}^k$ is $(\alpha_{k+1}, \beta_{k+1}, \gamma_{k+1})$-sharp by applying Proposition \ref{prop:sharp_comp} with $p_1 = \tfrac{k}{k+1}$ and $p_2 = \tfrac{1}{k+1}$, where
\begin{align}
\alpha_{k+1} &= \frac{k^2}{(k+1)^2} \alpha_k + \frac{1}{(k+1)^2} L \alpha + \frac{2k}{(k+1)^2} \beta + \frac{k^2}{L(k+1)^2} \gamma \\
\beta_{k+1} &= \frac{k}{k+1} \beta_{k} + \frac{1}{k+1} \beta + \frac{k}{L(k+1)} \gamma \\
\gamma_{k+1} &= \gamma_{k} + \frac{1}{L} \gamma.
\end{align}
which yields the induction.

Setting $k = L$, observe that $1 + 2 + \hdots + (L-1) = \tfrac{1}{2}L(L-1)$ and $1^2 + 2^2 + \hdots + (L-1)^2 = \tfrac{1}{6}L(L-1)(2L-1)$, giving
\begin{align}
    \alpha_L &= \alpha + \frac{L-1}{L} \beta + \frac{L(L-1)(2L-1)}{6 L^3} \gamma \leq \alpha + \beta + \tfrac{1}{3} \gamma \\
    \beta_L &= \beta + \frac{L-1}{2L} \gamma \leq \beta + \tfrac{1}{2}\gamma \\
    \gamma_L &= \gamma 
\end{align}
which proves the result.

\subsection*{\cref{thm:smoothness}: Smoothness in the modular norm}
\label[appendix]{app:smoothness}

To establish the first inequality, we start by applying the Gauss-Newton decomposition (\ref{eq:gauss-newton}) of the Hessian, followed by the fact that the error $\ell$ is $\sigma$-Lipschitz and $\tau$-smooth, followed by the well-normedness and $(\alpha,\beta,\gamma)$-sharpness of the module $\module$:
\begin{align}
    &\abs{\D \vw \diamond \nabla^2_{\vw \vw} \Loss \diamond \D \widetilde\vw} \\
    &\quad= \left|\Expect_{\vx,\vy\sim \data} \left[
    \nabla_{\vy} \ell \diamond \left( \D \vw \diamond \nabla^2_{\vw \vw} \module \diamond \D \widetilde\vw \right)
    + (\nabla_{\vw} \module \diamond \D \vw) \diamond \nabla^2_{\vy \vy} \ell \diamond (\nabla_{\vw} \module \diamond \D \widetilde\vw)\right]\right| \\
    &\quad\leq \Expect_{\vx,\vy\sim \data} \left[
    \sigma\, \norm{\D \vw \diamond \nabla^2_{\vw \vw} \module \diamond \D \widetilde\vw}_\outputs
    + \tau\, \norm{\nabla_{\vw} \module \diamond \D \vw}_\outputs \, \norm{\nabla_{\vw} \module \diamond \D \widetilde\vw}_\outputs\right]\\
    &\quad\leq
    (\sigma\alpha
    + \tau ) \, \norm{\Delta \vw}_\module \, \norm{\Delta \widetilde\vw}_\module.
\end{align}
The second inequality follows from the first via the fundamental theorem of calculus:
\begin{align}
    \norm{\nabla_\vw \el(\vw+\Delta \vw) - \nabla_\vw\el(\vw)}_\module^* &= \max_{\norm{\Delta\widetilde\vw}_\module=1}\left|\left[\nabla_\vw \el(\vw+\Delta \vw) - \nabla_\vw\el(\vw)\right] \diamond \Delta \widetilde\vw\right| \\
    &\leq \max_{\norm{\Delta\widetilde\vw}_\module=1} \smash{\int_0^1} \abs{ \Delta \vw \diamond \nabla^2_{\vw\vw}\el(\vw+t\Delta \vw) \diamond \Delta \widetilde \vw} \idiff{t}\\
    &\leq \max_{\norm{\Delta\widetilde\vw}_\module=1} (\sigma\alpha
    + \tau ) \, \norm{\Delta \vw}_\module \, \norm{\Delta \widetilde\vw}_\module \smash{\int_0^1} \diff{t} \\
    &= (\sigma\alpha
    + \tau ) \, \norm{\Delta \vw}_\module.
\end{align}
The third inequality follows from the second by again applying the fundamental theorem of calculus, followed by the Cauchy-Schwarz inequality:
\begin{align}
    &\left|\el(\vw+\Delta \vw) - \left[\el(\vw) + \nabla_\vw \el(\vw)\diamond \Delta \vw \right] \right|\\
    &\qquad= \left|\int_0^1 \left[\nabla_\vw \el (\vw + t\Delta \vw) - \nabla_\vw \el(\vw)\right] \diamond \Delta \vw \idiff{t}\right|\\
    &\qquad\leq \int_0^1 \norm{\nabla_\vw \el (\vw + t\Delta \vw) - \nabla_\vw \el(\vw)}_\module^* \,\norm{\Delta \vw}_\module \idiff{t} \\
    &\qquad\leq (\sigma\alpha + \tau) \, \norm{\Delta \vw}_\module^2 \int_0^1 t \idiff{t} \\
    &\qquad= \half \,(\sigma\alpha + \tau) \, \norm{\Delta \vw}_\module^2. 
\end{align}
This completes the proof.

\subsection*{\cref{prop:broadcast_sens}: Broadcast modules are well-normed}\label{sec:broadcast_sens_proof}

Suppose $\module$ is a module with inputs $\inputs$, outputs $\outputs$ and weights $\weights$, broadcast to take $\inputs^h$ to $\outputs^h$. We take norms on these spaces to be
    \begin{align}
        \norm{(\vx_1, \hdots, \vx_h)}_{\inputs^h} &= S* \left(\norm{\vx_1}_{\inputs}^p + \hdots + \norm{\vx_h}_{\inputs}^p \right)^{1/p} \\
        \norm{(\vy_1, \hdots, \vy_h)}_{\outputs^h} &= S* \left(\norm{\vy_1}_{\outputs}^p + \hdots + \norm{\vy_h}_{\outputs}^p \right)^{1/p}
    \end{align}
where $S = h^{-1/p}$ unless $\module$ is a bond module. Write $\mu = \module\lip$. Then, for perturbations in the weight direction, which only occur if $\module$ is not a bond module:
\begin{align}
    \norm{\nabla_{\vw} \module(\vw, \vx_1, \hdots, \vx_h) \diamond \D \vw}_{\outputs^h}
    &= \left( \frac{1}{h} \sum_j \norm{\nabla_{\vw} \module(\vw, \vx_j) \diamond \D \vw}_{\outputs}^p \right)^{1/p} \\
    &\leq \norm{\D \vw}_{\module} \quad \text{ applying well-normed-ness.}
\end{align}
For perturbations in the input direction, we have:
\begin{align}
    \norm{\nabla_{\vx_1, \hdots, \vx_h} \module \diamond (\D \vx_1, \hdots, \D \vx_h)}_{\outputs^h}
    &= S * \left(\sum_j \norm{\nabla_{\vx_j} \module \diamond \D \vx_j}^p_{\outputs}
    \right)^{1/p} \\
    &\leq S * \left(\sum_j \mu^p \norm{\D \vx_j}^p_{\outputs}\right)^{1/p} \\
    &= \mu * \norm{(\D \vx_1, \hdots, \D \vx_h)}_{\outputs^h}
\end{align}
which proves the proposition.

\subsection*{\cref{prop:attn-sens-sharpness}: Sensitivity of attention}\label{sec:attn_sensitivity}
We prove that the functional attention module $\funcattn$ of \cref{ex:attention} is well-normed and of unit sensitivity.

Recall we use the following norms on the inputs $\inputs = \R^{\ell \times d_Q} \times \R^{\ell \times d_Q} \times \R^{\ell \times d_V}$ and outputs $\outputs = \R^{\ell \times d_V}$:
\begin{equation}
    \norm{(\vq, \vk, \vv)}_{\inputs} = \norm{\vq}_{\inftyRMS} + \norm{\vk}_{\inftyRMS} + \norm{\vv}_{\inftyRMS}, \quad \norm{\vy}_{\outputs} = \norm{\vy}_{\inftyRMS}.
\end{equation}
We will also make use of the $L^{\infty}$-operator norm for $\ell \times \ell$ matrices, which we write as
\begin{equation}
    \norm{\mB}_{\inftyop} = \max_{i = 1, \hdots,\ell}\left( \sum_{j=1}^{\ell} \abs{\mB_{ij}}\right);
\end{equation}
observe that for $\mB \in \R^{\ell \times \ell}$ and $\vx \in \R^{\ell \times d}$ we have
\begin{equation}
    \norm{\mB \, \vx}_{\inftyRMS} \le \norm{\mB}_{\inftyop} \norm{\vx}_{\inftyRMS}.
\end{equation}
Writing $F = \funcattn\for$ for short, recall that
\begin{equation}
    F(\vq, \vk, \vv) = \softmax(\tfrac{1}{d_Q} \vq \vk^T + \mM) \vv
\end{equation}
where $\mM$ is the mask (our proof will apply equally for the standard causal mask and also the non-causal $\mM \equiv 0$).

We will prove that at any $(\vq, \vk, \vv)$ satisfying $\norm{\vq}_{\inftyRMS}, \norm{\vk}_{\inftyRMS}, \norm{\vv}_{\inftyRMS} \le 1$, for any $(\D \vq, \D \vk, \D \vv)$ we have
\begin{equation}
    \norm{\nabla F(\vq, \vk, \vv) \diamond (\D \vq, \D \vk, \D \vv)}_{\outputs} \le \norm{(\D \vq, \D \vk, \D \vv)}_{\inputs}.
\end{equation}
For short, write $\mA = \softmax(\tfrac{1}{d_Q} \vq \vk^T + \mM)$ for the attention matrix and its derivative as
\begin{equation}
    \D \mA = \nabla_{(\vq, \vk)} \softmax(\tfrac{1}{d_Q} \vq \vk^T + \mM) \diamond (\D \vq, \D \vk).
\end{equation}

Now, the derivative of $F$ splits into two terms
\begin{equation}
    \nabla F \diamond (\D \vq, \D \vk, \D \vv)
    = \mA (\D \vv)
    + (\D \mA) \vv.
\end{equation}

To complete the proof, we claim that
\begin{equation}
    \norm{\mA}_{\inftyop} = 1 \quad \text{and} \quad
    \norm{\D \mA}_{\inftyop} \le \norm{\D \vq}_{\inftyRMS} + \norm{\D \vk}_{\inftyRMS}.
\end{equation}
The calculation of the norm of $\mA$ follows by definition from its construction by softmax. For the calculation of the norm of $\D \mA$, a direct calculation yields that
\begin{equation}
    \D \mA_{ij} = \tfrac{1}{d_Q}\mA_{ij} \ip{\D \vq_{i}}{\vk_j - \Sigma_m \mA_{im} \vk_m} + \tfrac{1}{d_Q}\mA_{ij}\ip{\vq_{i}}{\D \vk_j - \Sigma_m \mA_{im} \D \vk_m}
\end{equation}
where we are writing $\vq_i = \vq_{i*}$ and so on.

Taking absolute values, applying the Cauchy-Schwarz inequality and summing over $j$ we have
\begin{align}
    \Sigma_j \abs{\D \mA_{ij}} \leq 
    &\norm{\D \vq_i}_{\RMS} \left(\Sigma_j \mA_{ij}\norm{\vk_j - \Sigma_m \mA_{im} \vk_m}_{\RMS}\right)\\
    &+ \norm{\vq_i}_{\RMS} \left(\Sigma_j \mA_{ij}\norm{\D \vk_j - \Sigma_m \mA_{im} \D \vk_m}_{\RMS}\right).
\end{align}
We now use the following inequality: given any non-negative reals $p_1, \hdots, p_{\ell}$ which sum to 1, and any vectors $\vx_1, \hdots, \vx_{\ell}$ in an inner product space with norm $\norm{\cdot}$, we have by Jensen's inequality
\begin{align}
    \Sigma_j p_j \norm{\vx_j - \Sigma_m p_m \vx_m}
    &\leq \left(\Sigma_j p_j \norm{\vx_j - \Sigma_m p_m \vx_m}^2\right)^{\tfrac{1}{2}} \\
    &= \left(
    \Sigma_j p_j \norm{\vx_j}^2 - \norm{\Sigma_j p_j \vx_j}^2
    \right)^{\tfrac{1}{2}} \\
    &\leq \left(\Sigma_j p_j \norm{\vx_j}^2 \right)^{\tfrac{1}{2}} \\
    &\leq \max_j \norm{\vx_j}.
\end{align}

Applying to the matrix $\D \mA$, we thus have
\begin{equation}
    \Sigma_j \abs{\mA_{ij}} \leq \norm{\D \vq_i}_{\RMS} \max_{j} \norm{\vk_j}_{\RMS} + \norm{\vq_i}_{\RMS} \max_j \norm{\D \vk_j}_{\RMS}.
\end{equation}
Taking the max over $i$, this shows the $L^{\infty}$-operator-norm of $\D \mA$ is at most
\begin{equation}
    \norm{\D \vq}_{\inftyRMS} \norm{\vk}_{\inftyRMS} + \norm{\vq}_{\inftyRMS} \norm{\D \vk}_{\inftyRMS}
\end{equation}
which, since $\norm{\vq}_{\inftyRMS}, \norm{\vk}_{\inftyRMS} \leq 1$, completes the proof.

\subsection*{\cref{prop:attn-sens-sharpness}: Sharpness of functional attention}

In this section, we estimate the second derivative of the forward function $F$ of functional attention at $(\vq, \vk, \vv)$ in perturbation directions $(\D \vq, \D \vk, \D \vv)$ and $(\D \widetilde{\vq}, \D \widetilde{\vk}, \D \widetilde{\vv})$:
\begin{equation}
\D^2 F := (\D \widetilde{\vq}, \D \widetilde{\vk}, \D \widetilde{\vq}) \diamond \nabla^2 F \diamond (\D \vq, \D \vk, \D \vv).
\end{equation}
We will prove that functional attention is $\gamma$-sharp where in fact $\gamma = 3$; this amounts to proving that
\begin{equation}
    \norm{\D^2 F} \leq 3 \norm{(\D \vq, \D \vk, \D \vv)}_{\inputs}
    \norm{(\D \widetilde{\vq}, \D \widetilde{\vk}, \D \widetilde{\vv})}_{\inputs}.
\end{equation}

We continue with all the notation of the previous section. Moreover, to simplify the calculation, we suppress all factors of $\tfrac{1}{d_Q}$ (indeed, one can absorb them as a rescaled inner product $\ip{\cdot}{\cdot}$). We also, in addition to the shorthand $\vx_i = \vx_{i*}$ for $\ell \times d$ matrices $\vx$, we adopt the shorthand for an $\ell \times \ell$ matrix $\mB$ and a $\ell \times d$ matrix $\vx$, and any $i, j = 1, \hdots, \ell$:
\begin{equation}
    [\mB, \vx]_{ij} := \vx_j - \Sigma_m \mB_{im} \vx_m.
\end{equation}
We note three crucial inequalities regarding $[\mB, \vx]$, \emph{for any $\ell \times \ell$ matrix $\mB$ with non-negative entries whose rows sum to 1, and $\ell \times d$ matrices $\vx, \vy$:}:
\begin{align}
\label{eq:bracket_1}    \Sigma_j \mB_{ij}\norm{[\mB, \vx]_{ij}} &\leq \max_j \norm{\vx_j}; \\
\label{eq:bracket_2}    \Sigma_j \mB_{ij}\norm{[\mB, \vx]_{ij}}^2 &\leq \max_j \norm{\vx_j}^2; \\
\label{eq:bracket_3}    \Sigma_j \mB_{ij}\norm{[\mB, \vx]_{ij}}\norm{[\mB, \vy]_{ij}} &\leq (\max_j \norm{\vx_j})(\max_j \norm{\vy_j}).
\end{align}
All three inequalities follow from standard expectation/variance inequalities for random variables on the finite set $\{1, \hdots, \ell\}$ with distributions given by $\mB_{i1}, \hdots, \mB_{i\ell}$.

With these conventions, the expression for $\D \mA$ is thus
\begin{equation}
    \D \mA_{ij} = \mA_{ij} \ip{\D \vq_i}{[\mA,\vk]_{ij}} + \mA_{ij} \ip{\vq_i}{[\mA,\D \vk]_{ij}}.
\end{equation}
Let us also write
\begin{align}
    \D \widetilde{\mA} :&= \nabla_{(\vq, \vk)} \softmax(\tfrac{1}{d_Q} \vq \vk^T + \mM) \diamond (\D \widetilde{\vq}, \D \widetilde{\vk}) \\
    \D \widetilde{\mA}_{ij} &= \mA_{ij} \ip{\D \widetilde{\vq}_i}{[\mA,\vk]_{ij}} + \mA_{ij} \ip{\vq_i}{[\mA,\D \widetilde{\vk}]_{ij}}.
\end{align}
as well as
\begin{equation}
    \D^2 \mA := (\D \widetilde{\vq}, \D \widetilde{\vk}) \diamond \nabla^2 F \diamond (\D \vq, \D \vk).
\end{equation}
In these terms, the second derivative $\D^2 F$ is just
\begin{equation}
    \D^2 F = (\D \widetilde{\mA})(\D \vv) + (\D \mA)(\D \widetilde{\vv}) + (\D^2 \mA)\vv.
\end{equation}
From the estimates of the previous section, we have
\begin{align}
\label{eq:DtilADv}    \norm{(\D \widetilde{\mA})(\D \vv)}_{\inftyRMS}
    &\leq
    (\norm{\D \widetilde{\vq}}_{\inftyRMS} + \norm{\D \widetilde{\vk}}_{\inftyRMS}) \norm{\D \vv}_{\inftyRMS} \\
\label{eq:DADtilv}    \norm{(\D {\mA})(\D \widetilde{\vv})}_{\inftyRMS}
    &\leq
    (\norm{\D {\vq}}_{\inftyRMS} + \norm{\D {\vk}}_{\inftyRMS}) \norm{\D \widetilde{\vv}}_{\inftyRMS}
\end{align}
so our task is to estimate the $L^{\infty}$-operator-norm of $\D^2 \mA$. Thus, we calculate $\D^2 \mA$:
\begin{align}
\label{eq:D2A_1}    \D^2 \mA_{ij} = \, &
    \mA_{ij}\ip{\D \vq_i}{[\mA, \D \widetilde{\vk}]_{ij}}\\
\label{eq:D2A_2}    &+ \mA_{ij}\ip{\D \widetilde{\vq}_i}{[\mA, \D \vk]_{ij}}\\
\label{eq:D2A_3}    &+ \D\widetilde{\mA}_{ij}\ip{\D \vq_i}{[\mA, \vk]_{ij}}\\
\label{eq:D2A_4}    &+ \D\widetilde{\mA}_{ij}\ip{\vq_i}{[\mA,\D \vk]_{ij}}\\
\label{eq:D2A_5}    &+ \mA_{ij}\ip{\D \vq_i}{-\Sigma_m (\D \widetilde{\mA})_{im}\vk_m}\\
\label{eq:D2A_6}    &+ \mA_{ij}\ip{\vq_i}{-\Sigma_m (\D \widetilde{\mA})_{im}\D \vk_m}
\end{align}
We estimate the $L^{\infty}$-operator-norm of these six terms one by one. The first (\ref{eq:D2A_1}), (\ref{eq:D2A_2}) are the simplest, using inequality (\ref{eq:bracket_1}):
\begin{align}
    \max_i \Sigma_j \abs{\mA_{ij}\ip{\D \vq_i}{[\mA, \D \widetilde{\vk}]_{ij}}}
    &\leq \max_i \Sigma_j \mA_{ij} \norm{\D \vq_i} \norm{[\mA, \D \widetilde{\vk}]_{ij}}\\
    &\leq \max_i \norm{\D \vq_i} \max_j \norm{\D \vk_j} \\
    &= \norm{\D \vq}_{\inftyRMS} \norm{\D \widetilde{\vk}}_{\inftyRMS} \\
    \max_i \Sigma_j \abs{\mA_{ij}\ip{\D \widetilde{\vq}_i}{[\mA, \D \vk]_{ij}}}
    &\leq \norm{\D \widetilde{\vq}}_{\inftyRMS} \norm{\D {\vk}}_{\inftyRMS} \quad \text{ likewise.}
\end{align}
For the term (\ref{eq:D2A_3}), we have
\begin{align}
    \D\widetilde{\mA}_{ij}\ip{\D \vq_i}{[\mA, \vk]_{ij}}
    &= \left(\mA_{ij} \ip{\D \widetilde{\vq}_i}{[\mA,\vk]_{ij}} + \mA_{ij} \ip{\vq_i}{[\mA,\D \widetilde{\vk}]_{ij}}\right)
    \ip{\D \vq_i}{[\mA, \vk]_{ij}}
\end{align}
Take absolute values, sum over $j$, and apply Cauchy-Schwarz and inequalities (\ref{eq:bracket_2}),(\ref{eq:bracket_3}):
\begin{align}
    \Sigma_j\abs{\D\widetilde{\mA}_{ij}\ip{\D \vq_i}{[\mA, \vk]_{ij}}}
    &\leq
    \Sigma_j \mA_{ij}
    \left(
    \norm{\D \vq_i}\norm{\D \widetilde{\vq}_i}\norm{[\mA, \vk]_{ij}}^2
    + \norm{\vq_i}\norm{\D \vq_i}\norm{[\mA, \vk]_{ij}}
    \norm{[\mA, \D\widetilde{\vk}]_{ij}}
    \right)\\
    &\leq
    \norm{\D \vq_i}\norm{\D \widetilde{\vq}_i}\max_j \norm{\vk_j}^2
    + \norm{\vq_i}\norm{\D \vq_i}(\max_j \norm{\vk_j})(\max_j \norm{\D \widetilde{\vk}_j}).
\end{align}
Taking the max over $i$ and applying $\norm{\vq}_{\inftyRMS}, \norm{\vk}_{\inftyRMS}, \norm{\vv}_{\inftyRMS} \le 1$:
\begin{equation}
    \max_i \Sigma_j\abs{\D\widetilde{\mA}_{ij}\ip{\D \vq_i}{[\mA, \vk]_{ij}}}
    \leq
    \norm{\D \vq}_{\inftyRMS} \norm{\D \widetilde{\vq}}_{\inftyRMS}
    + \norm{\D \vq}_{\inftyRMS}\norm{\D \widetilde{\vk}}_{\inftyRMS}.
\end{equation}
The term (\ref{eq:D2A_4}) is similar:
\begin{align}
    \max_i \Sigma_j \abs{\D\widetilde{\mA}_{ij}\ip{\vq_i}{[\mA,\D \vk]_{ij}}}
    &\leq
    \norm{\D \vk}_{\inftyRMS}\norm{\D \widetilde{\vq}}_{\inftyRMS} + \norm{\D \vk}_{\inftyRMS}\norm{\D \widetilde{\vk}}_{\inftyRMS}
\end{align}
For term (\ref{eq:D2A_5}), observe that
\begin{align}
    \max_i \norm{\Sigma_m (\D \widetilde{\mA})_{im}\vk_m}
    &\leq \norm{\D \widetilde{\mA}}_{\inftyop} \norm{\D \vk}_{\inftyRMS} \\
    &\leq \norm{\D \widetilde{\vq}}_{\inftyRMS} + \norm{\D \widetilde{\vk}}_{\inftyRMS}
\end{align}
and so by Cauchy-Schwarz and the fact that the rows of $\mA$ sum to 1:
\begin{align}
    \max_i \Sigma_j \abs{\mA_{ij}\ip{\D \vq_i}{-\Sigma_m (\D \widetilde{\mA})_{im}\vk_m}}
    &\leq \max_i \norm{\D \vq_i} \norm{\Sigma_m (\D \widetilde{\mA})_{im}\vk_m} \\
    &\leq \norm{\D \vq}_{\inftyRMS} \norm{\D \widetilde{\vq}}_{\inftyRMS} + \norm{\D \vq}_{\inftyRMS} \norm{\D \widetilde{\vk}}_{\inftyRMS}.
\end{align}
By a similar argument, for term (\ref{eq:D2A_6}) we have:
\begin{equation}
    \mA_{ij}\ip{\vq_i}{-\Sigma_m (\D \widetilde{\mA})_{im}\D \vk_m}
    \leq
    \norm{\D \vk}_{\inftyRMS} \norm{\D \widetilde{\vq}}_{\inftyRMS} + \norm{\D \vk}_{\inftyRMS} \norm{\D \widetilde{\vk}}_{\inftyRMS}
\end{equation}

Thus, we have an estimate on the $L^{\infty}$-operator-norm of $\D^2 \mA$:
\begin{equation}
\label{eq:D2A_est}    \norm{\D^2 \mA}_{\inftyop}
    \leq
    2 \norm{\D \vq}\norm{\D \widetilde{\vq}}
    + 3 \norm{\D \vq}\norm{\D \widetilde{\vk}}
    + 3 \norm{\D \vk}\norm{\D \widetilde{\vq}}
    + 2 \norm{\D \vk}\norm{\D \widetilde{\vk}}
\end{equation}
where all the norms on the right hand side are $\norm{\cdot}_{\inftyRMS}$.

Adding this together with (\ref{eq:DtilADv}) and (\ref{eq:DADtilv}), we obtain (all norms being $\norm{\cdot}_{\inftyRMS}$:
\begin{align}
    \norm{\D^2 F}
    \leq \, &
    2 \norm{\D \vq}\norm{\D \widetilde{\vq}}
    + 3 \norm{\D \vq}\norm{\D \widetilde{\vk}}
    + 3 \norm{\D \vk}\norm{\D \widetilde{\vq}}
    + 2 \norm{\D \vk}\norm{\D \widetilde{\vk}} \\
    &+ \norm{\D \vv}\norm{\D \widetilde{\vq}}
    + \norm{\D \vv}\norm{\D \widetilde{\vk}}
    + \norm{\D \vq}\norm{\D \widetilde{\vv}}
    + \norm{\D \vk}\norm{\D \widetilde{\vv}} \\
    \leq \, &
    3(\norm{\D \vq} + \norm{\D \vk} + \norm{\D \vv})
    (\norm{\D \widetilde{\vq}} + \norm{\D \widetilde{\vk}} + \norm{\D \widetilde{\vv}})
\end{align}
which is the desired result.

\subsection*{\cref{prop:sharp_comp}: Sharpness under composition}
Suppose $\module = \module_2 \circ \module_1$ where $\module_1, \module_2$ are well-normed modules on respectively $(\inputs_k, \outputs_k, \weights_k)$ and moreover $(\alpha_k, \beta_k, \gamma_k)$-sharp for $k=1,2$. If $p_k = \tfrac{\module_k\mass}{\module\mass}$ for $k=1,2$, note that by the definition of the modular norm on the composite $\module$, we have for any $\D \vw = (\D \vw_1, \D \vw_2) \in \weights_1 \times \weights_2$:
\begin{equation}
    \norm{\D \vw_1}_{\module_1} \le \tfrac{1}{\mu_2} p_1 \norm{\D \vw}_{\module}
    \quad \text{and} \quad
    \norm{\D \vw_2}_{\module_2} \le p_2 \norm{\D \vw}_{\module}.
\end{equation}
We must prove that $\module$ is $(\alpha, \beta, \gamma)$ sharp where:
\begin{align}
\label{eq:sharp_comp_alpha_app}    \alpha &= 
    \tfrac{1}{\mu_2} p_1^2 \alpha_1 + p_2^2 \alpha_2 + \tfrac{2}{\mu_2} p_1p_2 \beta_2 + \tfrac{1}{\mu_2^2} p_1^2 \gamma_2, \\
\label{eq:sharp_comp_beta_app}    \beta &= p_1 \beta_1 + \mu_1 p_2 \beta_2 + \tfrac{\mu_1}{\mu_2} p_1 \gamma_2, \\
\label{eq:sharp_comp_gamma_app}    \gamma &= \mu_2 \gamma_1 + \mu_1^2 \gamma_2.
\end{align}

Turning to the second derivative of $\module(\cdot, \cdot)$, we prove the first Inequality (\ref{eq:sharp_comp_alpha_app}). The Gauss-Newton decomposition (\ref{eq:gauss-newton}) for any $\D \vw = (\D \vw_1, \D \vw_2)$ and $\D \widetilde{\vw} = \D \widetilde{\vw}_1, \D \widetilde{\vw}_2$ yields
\begin{align}
 \label{eq:GN_comp_term1}   \D \vw \diamond \nabla^2 \module \diamond \D \widetilde{\vw}
    = &\nabla \module_2 \diamond (\D \vw_1 \diamond \nabla^2 \module_1 \diamond \D \widetilde{\vw}_1) \\
\label{eq:GN_comp_term2}    &+ (\D \vw_2, \nabla \module_1 \diamond \D \vw_1) \diamond \nabla^2 \module_2 \diamond (\D \widetilde{\vw}_2, \nabla \module_1 \diamond \D \widetilde{\vw}_1)
\end{align}
Applying the well-normed and sharpness inequalities, the norm of the first (\ref{eq:GN_comp_term1}) of these terms is bounded by
\begin{align}
    \mu_2 \norm{\D \vw_1 \diamond \nabla^2 \module_1 \diamond \D \widetilde{\vw}_1}_{\outputs_1}
    &\leq \mu_2 \alpha_1 \norm{\D \vw_1}_{\module_1} \norm{\D \widetilde{\vw}_1}_{\module_1}\\
    &\leq \tfrac{1}{\mu_2}p_1^2 \alpha_1 \norm{\D \vw}_{\module}\norm{\D \widetilde{\vw}}_{\module}.
\end{align}
The second term (\ref{eq:GN_comp_term2}) breaks into four separate terms:
\begin{align}
    &\D \vw_2 \diamond \nabla_{\vw \vw}^2 \module_2 \diamond \D \widetilde{\vw}_2 \\
    + &(\nabla \module_1 \diamond \D \vw_1) \diamond \nabla_{\vx \vw}^2 \module_2 \diamond \D \widetilde{\vw_2} \\
    + & \D \vw_2 \diamond \nabla_{\vw \vx}^2 \module_2 \diamond (\nabla \module_1 \diamond \D \widetilde{\vw}_1) \\
    + & (\nabla \module_1 \diamond \D \vw_1) \diamond \nabla_{\vx \vx}^2 \module_2 \diamond (\nabla \module_1 \diamond \D \widetilde{\vw}_1).
\end{align}
In particular, applying the well-normed and sharpness inequalities, this is bounded by
\begin{align}
    &\alpha_2 \norm{\D \vw_2}_{\module_2}\norm{\D \widetilde{\vw}_2}_{\module_2}
    \\+& \beta_2 \norm{\D \vw_1}_{\module_1} \norm{\D \widetilde{\vw}_2}_{\module_2}
    \\+& \beta_2 \norm{\D \vw_2}_{\module_2} \norm{\D \widetilde{\vw}_1}_{\module_1}
    \\+& \gamma_2 \norm{\D \vw_1}_{\module_1} \norm{\D \widetilde{\vw}_1}_{\module_1},
\end{align}
which is less than
\begin{equation}
    \left(p_2^2 \alpha_2 + \tfrac{2}{\mu_2} p_1 p_2 \beta_2 + \tfrac{1}{\mu_2^2} p_1^2 \gamma_2 \right) \norm{\D \vw}_{\module}\norm{\D \vw}_{\module}
\end{equation}
which completes the proof of Inequality (\ref{eq:sharp_comp_alpha}).

Inequalities (\ref{eq:sharp_comp_beta_app}) and (\ref{eq:sharp_comp_gamma_app}) are simpler. For the first of these, note we have
\begin{align}
\label{eq:GN_comp_term3}   \D \vw \diamond \nabla^2 \module \diamond \D \vx
    =& \nabla \module_2 \diamond (\D \vw_1 \diamond \nabla^2 \module_1 \diamond \D \vx) \\
\label{eq:GN_comp_term4}    &+ (\D \vw_2, \nabla \module_1 \diamond \D \vw_1) \diamond \nabla^2 \module_2 \diamond (\nabla \module_1 \diamond \D \vx).
\end{align}
Term (\ref{eq:GN_comp_term3}) is bounded by
\begin{align}
    \mu_2 \norm{\D \vw_1 \diamond \nabla^2 \module_1 \diamond \D \vx}_{\outputs_1}
    &\leq \mu_2 \beta_1 \norm{\D \vw_1}_{\module_1} \norm{\D \vx}_{\inputs_1} \\
    &\leq p_1 \beta_1 \norm{\D \vw}_{\module} \norm{\D \vx}_{\inputs_1}
\end{align}
Term (\ref{eq:GN_comp_term4}) breaks into two separate terms
\begin{equation}
    \D \vw_2 \diamond \nabla^2_{\vw \vx} \module_2 \diamond (\nabla \module_1 \diamond \D \vx)
    + (\nabla \module_1 \diamond \D \vw_1) \diamond \nabla^2_{\vx \vx} \module_2 \diamond (\nabla \module_1 \diamond \D \vx).
\end{equation}
In particular this is bounded by
\begin{equation}
    \beta_2 \norm{\D \vw_2}_{\module_2} \mu_1 \norm{\D \vx}_{\inputs_1}
    + \gamma_2 \norm{\D \vw_1}_{\module_1} \mu_1 \norm{\D \vx}_{\inputs_1}
    \leq
    \left(\mu_1 p_2 \beta_2 + \tfrac{\mu_1}{\mu_2} p_1 \gamma_2
    \right)\norm{\D \vw}_{\module}\norm{\D \vx}_{\inputs_1}
\end{equation}
which completes the proof of Inequality (\ref{eq:sharp_comp_beta_app}).

Finally, for (\ref{eq:sharp_comp_gamma_app}), we have
\begin{align}
\label{eq:GN_comp_term5}    \D \vx \diamond \nabla^2 \module \diamond \D \widetilde{\vx}
    =& \nabla \module_2 \diamond (\D \vx \diamond \nabla^2 \module_1 \diamond \D \widetilde{\vx}) \\
\label{eq:GN_comp_term6}    &+ (\nabla \module_1 \diamond \D \vx) \diamond \D^2 \module_2 \diamond (\nabla \module_1 \diamond \D \widetilde{\vx}).
\end{align}

Term (\ref{eq:GN_comp_term5}) is bounded by
\begin{equation}
    \mu_2 \norm{\D \vx \diamond \nabla^2 \module_1 \diamond \D \widetilde{\vx}}_{\outputs_1} \le \mu_2 \gamma_1 \norm{\D \vx}_{\inputs_1} \norm{\D \widetilde{\vx}}_{\inputs_1}
\end{equation}
while Term (\ref{eq:GN_comp_term6}) is bounded by
\begin{equation}
    \gamma_2 \norm{\nabla \module_1 \diamond \D \vx}_{\inputs_2} \norm{\nabla \module_1 \diamond \D \widetilde{\vx}}_{\inputs_2}
    \leq
    \mu_1^2 \gamma_2 \norm{\D \vx}_{\inputs_1} \norm{\D \widetilde{\vx}}_{\inputs_1}
\end{equation}
which together give Inequality (\ref{eq:sharp_comp_gamma_app}).

\subsection*{\cref{prop:sharp_concat}: Sharpness under concatenation}

Suppose $\module = (\module_1, \module_2)$ where $\module_1, \module_2$ are well-normed modules on respectively $(\inputs_k, \outputs_k, \weights_k)$ and moreover $(\alpha_k, \beta_k, \gamma_k)$-sharp for $k=1,2$. If $p_k = \tfrac{\module_k\mass}{\module\mass}$ for $k=1,2$, as in the previous proof we have for any $\D \vw = (\D \vw_1, \D \vw_2) \in \weights_1 \times \weights_2$:
\begin{equation}
    \norm{\D \vw_1}_{\module_1} \le \tfrac{1}{\mu_2} p_1 \norm{\D \vw}_{\module}
    \quad \text{and} \quad
    \norm{\D \vw_2}_{\module_2} \le p_2 \norm{\D \vw}_{\module}.
\end{equation}
We must prove that $\module$ is $(\alpha,\beta,\gamma)$-sharp where
\begin{align}
    \alpha &= 
    p_1^2 \alpha_1 + p_2^2 \alpha_2, \\
    \beta &= p_1 \beta_1 + p_2 \beta_2, \\
    \gamma &= \gamma_1 + \gamma_2.
\end{align}

Now, for the first of these identities, we have for $\D \vw = (\D \vw_1, \D \vw_2)$ and $\D \widetilde{\vw} = (\D \widetilde{\vw}_1, \D \widetilde{\vw}_2)$:
\begin{align}
    \norm{\D \vw \diamond \nabla^2 \module \diamond \D \widetilde{\vw}}_{\outputs_1 \times \outputs_2}
    &= \norm{(\D \vw_1 \diamond \nabla^2 \module_1 \diamond \D \widetilde{\vw}_1, \D \vw_2 \diamond \nabla^2 \module_2 \diamond \D \widetilde{\vw}_2)}_{\outputs_1 \times \outputs_2} \\
    &= \norm{\D \vw_1 \diamond \nabla^2 \module_1 \diamond \D \widetilde{\vw}_1}_{\outputs_1} + \norm{\D \vw_2 \diamond \nabla^2 \module_2 \diamond \D \widetilde{\vw}_2)}_{\outputs_2} \\
    &\leq \alpha_1 \norm{\D \vw_1}_{\module_1}^2 + \alpha_2 \norm{\D \vw_2}_{\module_2}^2 \\
    &\leq (p_1^2 \alpha_1 + p_2^2 \alpha_2) \, \norm{\D \vw}_{\module}^2
\end{align}
which shows $\alpha = p_1^2 \alpha_1 + p_2^2 \alpha_2$. The expressions for $\beta, \gamma$ follow similarly.

\end{document}